%% file: main.tex
\documentclass[sigconf,natbib=true]{acmart}
\AtBeginDocument{%
  \providecommand\BibTeX{{%
    \normalfont B\kern-0.5em{\scshape i\kern-0.25em b}\kern-0.8em\TeX}}}

\copyrightyear{2022}
\acmYear{2022}
\setcopyright{rightsretained}

\acmConference[CIKM '22]{Proceedings of the 31st ACM International
Conference on Information and Knowledge Management}{October
17--21, 2022}{Atlanta, GA, USA}
\acmBooktitle{Proceedings of the 31st ACM International Conference on
Information and Knowledge Management (CIKM '22), October 17--21,
2022, Atlanta, GA, USA}
\acmDOI{10.1145/3511808.3557459}
\acmISBN{978-1-4503-9236-5/22/10}

\usepackage{etoolbox}
\makeatletter
\patchcmd{\maketitle}{\@copyrightpermission}{
   \begin{minipage}{0.3\columnwidth}
     \href{https://creativecommons.org/licenses/by/4.0/}{\includegraphics[width=0.90\textwidth]{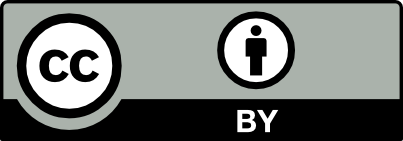}}
   \end{minipage}\hfill
   \begin{minipage}{0.7\columnwidth}
     \href{https://creativecommons.org/licenses/by/4.0/}{This work is licensed under a Creative Commons Attribution International 4.0 License.}
   \end{minipage}

   \vspace{5pt}
}{}{}

\makeatother

\usepackage{latexsym}
\usepackage{xspace}
\usepackage{booktabs}
\usepackage{enumitem}
\usepackage{multirow}
\usepackage{makecell}
\usepackage{color}
\usepackage{microtype}
\usepackage{amsmath}
\usepackage{graphicx}
\usepackage{xcolor}
\usepackage{subcaption}
\usepackage{caption}
\usepackage{microtype}
\newcommand{\our}{\mbox{\textsc{SPOT}}\xspace}

\renewcommand{\shortauthors}{Jiacheng Li et al.}

\begin{document}

%%
%% The "title" command has an optional parameter,
%% allowing the author to define a "short title" to be used in page headers.
%\title{Representing Knowledge by Spans: A Knowledge-Enhanced Model for Information Extraction}
\title{SPOT: Knowledge-Enhanced Language Representations for Information Extraction}
%%
%% The "author" command and its associated commands are used to define
%% the authors and their affiliations.
%% Of note is the shared affiliation of the first two authors, and the
%% "authornote" and "authornotemark" commands
%% used to denote shared contribution to the research.
\author{Jiacheng Li}
\affiliation{\country{University of California, San Diego}}
\email{j9li@eng.ucsd.edu}

\author{Yannis Katsis}
\affiliation{\country{IBM Research}}
\email{yannis.katsis@ibm.com}

\author{Tyler Baldwin}
\affiliation{\country{IBM Research}}
\email{tbaldwin@us.ibm.com}

\author{Ho-Cheol Kim}
\affiliation{\country{IBM Research}}
\email{hckim@us.ibm.com}

\author{Andrew Bartko}
\affiliation{\country{University of California, San Diego}}
\email{abartko@eng.ucsd.edu}

\author{Julian McAuley}
\affiliation{\country{University of California, San Diego}}
\email{jmcauley@eng.ucsd.edu}

\author{Chun-Nan Hsu}
\affiliation{\country{University of California, San Diego}}
\email{chunnan@ucsd.edu}

%\pagestyle{plain}
% CIKM page limit: 9 + ref

%%
%32% The abstract is a short summary of the work to be presented in the
%% article.
\begin{abstract}
\input{0_abstract}
\end{abstract}

%%
%% The code below is generated by the tool at http://dl.acm.org/ccs.cfm.
%% Please copy and paste the code instead of the example below.
%%


%%
%% Keywords. The author(s) should pick words that accurately describe
%% the work being presented. Separate the keywords with commas.

%%
%% This command processes the author and affiliation and title
%% information and builds the first part of the formatted document.
\maketitle

\section{Introduction}
\input{1_introduction}
\section{Related Works}
\input{2_related_works}

\section{Method}
\input{3_method}

\section{Experiments}
\label{sec:exp}
\input{4_experiment}

\section{Conclusion}
\input{5_conclusion}

\begin{acks}
This work is supported by IBM Research AI through the AI Horizons Network.
\end{acks}

%%
%% The next two lines define the bibliography style to be used, and
%% the bibliography file.
\bibliographystyle{ACM-Reference-Format}
\bibliography{sample-base}

%%
%% If your work has an appendix, this is the place to put it.
% \appendix
% \input{6_appendix}

\end{document}

%% file: 0_abstract.tex
Knowledge-enhanced pre-trained models for language representation have been shown to be more effective in 
knowledge base construction tasks (i.e.,~relation extraction)
than language models such as BERT. These knowledge-enhanced language models incorporate knowledge into pre-training to generate representations of entities or relationships. However, existing methods typically represent each entity with a separate embedding. As a result, these methods struggle to represent out-of-vocabulary entities and a large amount of parameters, on top of their underlying token models (i.e.,~the transformer), must be used and the number of entities that can be handled is limited in practice due to memory constraints. Moreover, existing models still struggle to represent entities and relationships simultaneously. To address these problems, we propose a new pre-trained model that learns representations of both entities and relationships from token spans and span pairs in the text respectively. By encoding spans efficiently with span modules, our model can represent both entities and their relationships but requires fewer parameters than existing models. We pre-trained our model with the knowledge graph extracted from Wikipedia and test it on a broad range of supervised and unsupervised information extraction tasks. Results show that our model learns better representations for both entities and relationships than baselines, while in supervised settings, fine-tuning our model outperforms RoBERTa consistently and achieves competitive results on information extraction tasks.

% 1. the second paragraph of the introduction
% 2. past tense in the experiments
% 3. our model instead of 'we' in the method section (passive mode)

%% file: 1_introduction.tex
Language %representation 
models pre-trained on a large amount of text such as BERT~\cite{Devlin2019BERTPO}, GPT~\cite{Radford2018ImprovingLU}, and BART~\cite{Lewis2020BARTDS} have achieved 
the
state-of-the-art on a wide variety of natural language processing (NLP) tasks. The % designed 
various % effective 
self-supervised learning objectives, such as masked language modeling~\cite{Devlin2019BERTPO}, enable language models to effectively
learn syntactic and semantic information from large text corpora without annotations. 
The learned
information can then be transferred 
to
downstream tasks, such as text classification~\cite{Wang2018GLUEAM}, named entity recognition~\cite{Sang2003IntroductionTT}, and question answering~\cite{Rajpurkar2016SQuAD10}.

Knowledge graphs (KG) provide a rich source of knowledge that can benefit information extraction tasks such as named entity recognition~\cite{Jiao2020SeNsERLC}, relation extraction~\cite{zhang2017tacred} and event extraction~\cite{Trieu2020DeepEventMineEN}.
Despite pre-trained models achieving success on a broad range of tasks, recent studies~\cite{Poerner2019BERTIN} % have 
suggested that language models pre-trained with unstructured text struggled to generate vectorized representations (i.e., embeddings) of entities and relationships and 
injecting prior knowledge from KG
% knowledge graphs % which are important 
to language models were attempted. 

\begin{table}[t]
\centering
\small
\begin{tabular}{lrcc}
\toprule
          & \# params. & Entity & Relation \\
Methods         & (Millions) & representation & representation \\ \midrule
ERNIE~\cite{Zhang2019ERNIEEL}  & 483.9 & Yes & No            \\
KnowBERT~\cite{Peters2019KnowledgeEC}  & 413.3 & Yes & No \\
LUKE~\cite{Yamada2020LUKEDC}      & 128 & Yes & No  \\
K-Adapter~\cite{Wang2020KAdapterIK} & 42 & No & No \\ 
SPOT (Ours)      & 21   & Yes         & Yes           \\ \bottomrule
\end{tabular}
\caption{Comparison of current knowledge-enhanced language
models based on the number of additional parameters (in millions) and whether they learn entity or relation representations.}
\label{tab:params}
\vspace{-8mm}
\end{table}

% Recently, some 
Many attempts have been made to inject knowledge into pre-trained language models~\cite{Zhang2019ERNIEEL,Lauscher2019InformingUP, Peters2019KnowledgeEC, Levine2020SenseBERTDS, Xiong2020PretrainedEW, Wang2020KAdapterIK, Yamada2020LUKEDC}. These previous works typically used separate embeddings for knowledge (i.e.,~entities and relationships in a KG) during pre-training
or fine-tuning on % the 
downstream tasks. For example, ERNIE~\cite{Zhang2019ERNIEEL} first applied TransE~\cite{Bordes2013TranslatingEF} on KG to obtain entity embeddings and then infused entities' embeddings into the language model. LUKE~\cite{Yamada2020LUKEDC} trained entity embeddings with token embeddings together in a transformer~\cite{Vaswani2017AttentionIA} by predicting masked tokens or entities. 
However, separate embeddings occupy extra memory and limit the number of knowledge entries that the model can handle. As is shown in Table~\ref{tab:params}, ERNIE and KnowBERT~\cite{Peters2019KnowledgeEC} need more than $400$ million parameters for entity embeddings. After restricting the number of entities, LUKE needs $128$ million parameters for $500$ thousand entities. K-Adapter~\cite{Wang2020KAdapterIK} % has 
uses only $42$ million additional parameters~\footnote{Additional parameters do not include parameters used in the
backbone model (e.g., BERT) for fair comparison.}, but cannot generate entity representations. Hence, to represent entities, %by 
previous works 
required
a large number of additional parameters. Additionally, the separate embedding tables limit the ability to representing out-of-vocabulary entities. For example, LUKE was pre-trained on the wikipedia article for entity representation but it struggles to represent biomedical entities in our entity clustering experiments (Section~\ref{sec:entity_cls}).
Furthermore, existing methods focused only on entity representations and lack the capacity to represent relationships to support downstream tasks for knowledge base construction such as relation extraction. As shown in our results (Figures~
\ref{fig:relation_repr}), simple methods based on entity represenations (e.g.~by concatenation) struggled to
produce informative relation representations. 
% from entity representations. 
ERICA~\cite{Qin2021ERICAIE} proposed to apply contrastive learning on entity and relation representations, but its use of simple average pooling on token representations cannot represent entities effectively especially for tasks that need entity boundaries such as joint entity and relation extraction.
As a result, 
we still need an effective pre-trained model that can incorporate external knowledge graphs into language modeling, and simultaneously learn representations of both entities and relationships without adding hundreds of millions of parameters to the model to support knowledge base construction tasks.

In this paper, we presented 
\our (SPan-based knOwledge Transformer)~\footnote{The codebase will be released upon acceptance.}, 
a span-based language model
that can be pre-trained 
to represent knowledge (i.e.,~entities and relationships) by encoding spans from 
input text. Intuitively, words, entities and relationships are hierarchical (i.e.,~entities contains words, relationships contains entities), it is possible to learn representations of entities from words and representations of relationships from entities. Hence, to incorporate knowledge from a KG, we can represent entities and relations by token spans and span pairs, respectively. There are two advantages of encoding spans:
(1) The pre-trained span encoder and pair encoder can effectively represent knowledge with much less parameters than works with separate embeddings. Different from previous works that learn an embedding for each entity, \our learns transformation from the language model's token representations to entity representations. Therefore, as shown in Table~\ref{tab:params}, \our % has 
needs
only $21$ million parameters (16 million for entities and 5 million for relationships) to incorporate knowledge into language model.
(2) Based on a span encoder and span pair encoder, \our learns both entity representations and relation representations in a unified way. Specifically, the span encoder learns entity representations from tokens and the span pair encoder learns relation representations from entities. In this hierarchical way, \our can effectively represent entities which do not appear in pre-training because our entity representations consist of token representations. With spans, \our incorporates knowledge into language modeling and also generates representations of both entities and relations. 

Specifically, to inject knowledge into a span-based model, we first
designed an effective model architecture to encode spans and span pairs.
Then, we adopted three pre-training tasks that jointly learn knowledge at three different levels, i.e., token, entity and relation. Specifically, at the token level, we masked random tokens as 
% previous 
in the
masked language model (MLM)
used in the pre-training of BERT~\cite{Devlin2019BERTPO}.
% and mask tokens % from part of within entities. 
% For 
At the entity and relation levels, the pre-training tasks are to predict entities and relations based on representations generated by the
span encoder and span pair encoder, respectively. In this way, \our learns to infer entities and relations from both entity mentions and 
their
contexts in the training text during the pre-training.

We pre-trained \our on Wikipedia 
text 
and used Wikidata as 
the 
corresponding knowledge graph.
After pre-training, we conducted extensive experiments on five benchmark datasets across entity and relationship clustering tasks and three supervised information extraction tasks,
namely
joint entity and relation extraction, entity typing and relation extraction. Experiments showed that 
fine-tuning
our pre-trained model consistently outperformed RoBERTa~\cite{Liu2019RoBERTaAR}, and achieved new state-of-the-art performance on four datasets.
In addition,
we compared our model with RoBERTa and LUKE by visualization to assess how well our pre-trained model learns representations of entities and relationships from pre-training.
Results indicated that our pre-trained model 
learned meaningful representations for both entities and relationships to support various knowledge base construction tasks.

Our contributions are in three folds:
\begin{itemize}
    \item We proposed to apply spans to effectively inject knowledge into a language model and generate highly informative entity and relationship representations.
    \item We designed a novel pre-training objective and trained a span-based framework with the Wikipedia dataset.
    \item Extensive experiments were conducted and showed that \our outperformed other knowledge-enhanced language model on information extraction tasks and generated superior knowledge representations.
\end{itemize}

%% file: 2_related_works.tex
\subsection{Joint Entity and Relation Extraction}
Since entity detection and relation classification are two essential tasks for knowledge base construction, numerous works~\cite{Shang2018LearningNE, Li2021Weakly, BaldiniSoares2019MatchingTB} were proposed on the two tasks. Because entity and their relationship recognition can benefit from exploiting interrelated signals, many models for joint detection of entities and relations were proposed recently. Most approaches used special tagging scheme for this tasks. \citet{Miwa2014ModelingJE} modeled joint entity and relation extraction as a table-filling problem, where each cell of the table corresponded to a word pair of the sentence. The BILOU tags were filled into the diagonal of the table and relation types were predicted in the off-diagonal cells. Similar to \citet{Miwa2014ModelingJE}, \citet{Gupta2016TableFM} also formulated joint learning as table filling problem but used a bidirectional recurrent neural network to label each word pair. Different from the previous works with special tagging scheme, \citet{Miwa2016EndtoEndRE} first applied a bidirectional sequential LSTM to tag the entities with BILOU scheme, and then a tree-structured RNN encoded the dependency tree between each entity pair to predict the relation type. IO-based tagging models cannot assign multiple tags on one token which limited the ability to recognizing multiple entities containing the common tokens. Hence, span-based approaches which performed an exhaustive search over all spans in a sentence were investigated and these approaches can cover overlapping entities. \citet{Dixit2019SpanLevelMF} and \citet{Luan2018MultiTaskIO} used span representations derived from a BiLSTM over concatenated ELMo~\cite{Peters2018DeepCW} word and character embeddings. These representations were then used across the downstream tasks including entity recognition and relation classification. DyGIE~\cite{Luan2019AGF} followed \citet{Luan2018MultiTaskIO} and added a graph propagation step to capture the interactions among spans. With emerging of contextualized span representations, further improvements were observed. DyGIE++~\cite{Wadden2019EntityRA} replaced the BiLSTM encoder with BERT. Other span applications included semantic role labeling~\cite{Ouchi2018ASS} and co-reference resolution~\cite{Lee2017EndtoendNC}. In this work, we used spans to incorporate %heterogeneous 
entities and relationships from a knowledge graph into a language model. Due to the flexibility of spans, our model can output knowledge-aware token representations and knowledge representations simultaneously to support joint extraction of entities and relationships.

\subsection{Pre-trained Language Representations} 
Early research on language representations focused on static unsupervised word representations such as Word2Vec~\cite{Mikolov2013EfficientEO} and GloVe~\cite{Pen2014Glove}. The basic idea was to leverage co-occurrences to learn latent word vectors that approximately reflected word semantics.
\citet{Dai2015SemisupervisedSL} and \citet{Howard2018UniversalLM} first pre-trained universal language representations on unlabeled text, and applied task-specific fine-tuning on downstream tasks.
Recent studies~\cite{Peters2018DeepCW, McCann2017LearnedIT} showed that contextual language representations were more powerful than static word embeddings %given 
because words could have different meanings in different contexts. 
Advances of transformer-based language models~\cite{Devlin2019BERTPO, Liu2019RoBERTaAR, Radford2018ImprovingLU} continued to improve contextual word representations with more efficient large-scale models and novel pre-training objectives. These approaches demonstrated their superiority in various downstream NLP tasks. Hence, many language model extensions had been proposed to further improve the performance. SpanBERT~\cite{Joshi2020SpanBERTIP} extended BERT by masking contiguous random spans rather than random tokens. Different from SpanBERT which predicted tokens by spans to obtain token representations, in our model, we apply a hierarchical structure (i.e.,~tokens, spans, span pairs) to represent tokens, entities and relationships in sentences. \citet{Song2019MASSMS} and \citet{Raffel2020ExploringTL} explored the impacts of various model architectures and % some works
others~\cite{Raffel2020ExploringTL, Lan2020ALBERTAL, Fedus2021SwitchTS} explored % larger 
enlarged model sizes to % achieve better 
improve general language understanding ability. 
MASS~\cite{Song2019MASSMS} and BART~\cite{Lewis2020BARTDS} extended the transformer encoder to the sequence-to-sequence architecture for pre-training. 
%To include heterogeneous data, 
% Some works improved natural language understanding by 
Multilingual learning~\cite{Lample2019CrosslingualLM, Tan2019LXMERTLC, Kondratyuk201975L1} and multi-modal learning~\cite{Lu2019ViLBERTPT, Sun2019VideoBERTAJ, Su2020VLBERTPO} were introduced to the pre-training.
Although these pre-trained language models achieved success in various NLP tasks, they still focused on token-level representations but ignored the entities and their relations existing in the sentences, which are crucial for downstream tasks related to knowledge extraction and management. Our model is also based on the transformer but we focused on injecting knowledge from knowledge graphs into pre-trained models built on spans. Compared to the aforementioned pre-trained language models, our model is able to incorporate knowledge into representation learning and generate highly informative entity and relation representations for downstream tasks.
% at the same time.

\subsection{Knowledge-Enhanced Language Representations} 
Contextual pre-trained language models provided word representations with rich semantic and syntactic information, but these models still struggled to represent knowledge (i.e.,~entities and relationships), 
% , which is important for some knowledge-driven tasks (e.g.,~information extraction tasks). 
% Some e
Efforts were made to improve learning of the representations of entities and relationships by injecting knowledge graphs into language models. % These works 
Early attempts enforced language models to memorize information about entities in a knowledge graph 
% and proposed 
with novel pre-training objectives. For example, ERNIE~\cite{Zhang2019ERNIEEL} aligned entities from Wikipedia sentences with fact triples in WikiData %. They trained entity embeddings using fact triples from WikiData 
via TransE~\cite{Bordes2013TranslatingEF}. % In the pre-training process, t
Their pre-training objective was to predict correct token-entity alignment from token and entity embeddings. KnowBERT~\cite{Peters2019KnowledgeEC} incorporated knowledge bases into BERT using knowledge attention and re-contextualization. Both ERNIE and KnowBERT enhanced language modeling by static entity embeddings separately learned from KGs. WKLM~\cite{Xiong2020PretrainedEW} replaced entity mentions in the original document and the model was trained to distinguish the correct entity mention from randomly chosen ones. KEPLER~\cite{Wang2021KEPLERAU} encoded textual entity descriptions with pre-trained language models as entity embeddings, and then jointly optimized the knowledge embedding and language modeling objectives. 
GreaseLM~\cite{Zhang2022GreaseLMGR} improved question answering by fusing encoded representations from pre-trained language models and graph neural networks over multiple layers of modality interaction (i.e.,~graphs and text) operations to obtain information from both modalities. Hence, this method allowed language context representations to be grounded by structured world knowledge. 
LUKE~\cite{Yamada2020LUKEDC} applied trainable entity embeddings and an improved transformer architecture to learn 
% train 
word and entity % embeddings 
representations together. 
Another line of works~\cite{Ye2020CoreferentialRL, Kong2020AMI, Qin2021ERICAIE} modeled the intrinsic relational facts in text data, making it easy to 
%understand 
represent out-of-KG knowledge in % the 
downstream tasks. Some works~\cite{BaldiniSoares2019MatchingTB, Peng2020LearningFC} focused only on relationships and learned to extract relations
%-aware semantics 
from text by comparing the sentences that share the same entity pair or distantly supervised relation in KG.

Unlike the methods mentioned above, we used spans and span pairs to represent entities and relationships. After the pre-training, our knowledge enhanced language model can incorporate knowledge of KG into token representations and directly output meaningful % knowledge 
representations of entities and relationships from given spans and span pairs without fine-tuning. Compared to previous works that used separate embedding tables for entities~\cite{Zhang2019ERNIEEL, Peters2019KnowledgeEC, Yamada2020LUKEDC} or encoded entities using multiple language models~\cite{Wang2021KEPLERAU}, our model is 
%superior 
novel in terms of leveraging span-based representations to achieve simplicity, efficiency and effectiveness. % with span representations.

%% file: 3_method.tex
\begin{figure*}[t]
\centering
 \includegraphics[width=0.9\textwidth]{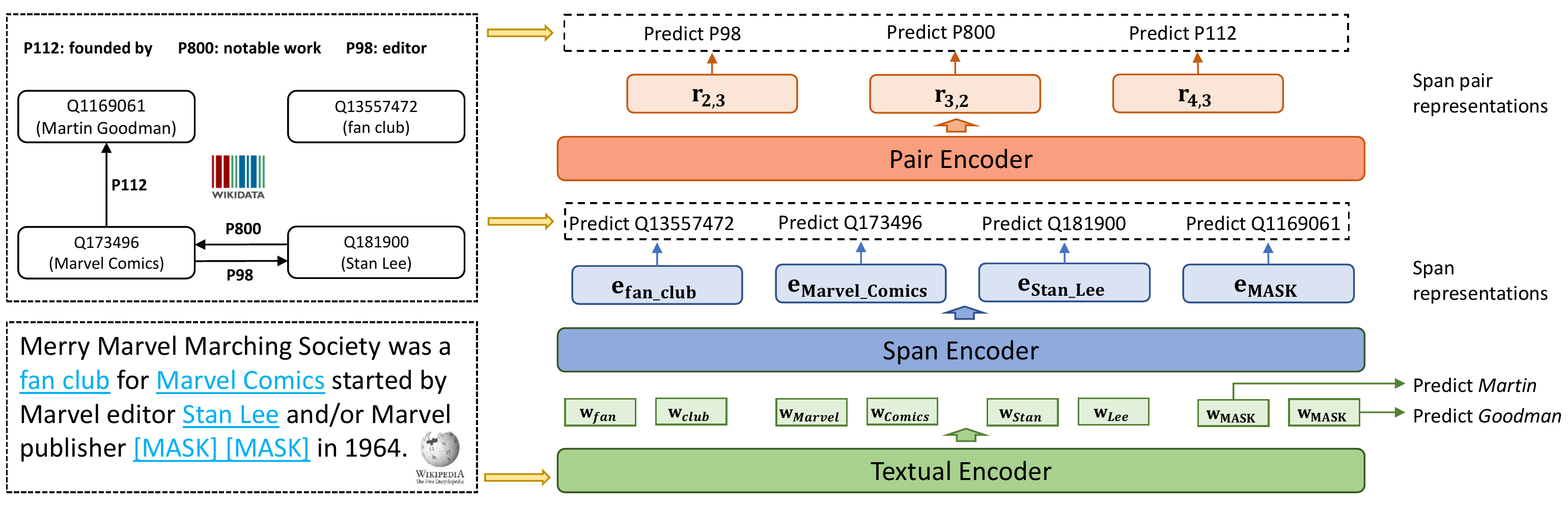}
\vspace{-4mm}
\caption{
Overview of \our~using an input sentence from pre-training dataset.
}
\vspace{-3mm}
\label{fig:overview}
\end{figure*}

This section presents the overall framework of \our and its detailed implementation, including the model architecture (Section~\ref{sec:model_arch}) and the novel pre-training task designed for incorporating the knowledge of entities and relationships from a knowledge graph (KG) (Section~\ref{sec:pre_training}).
\subsection{Notation}
Given a text corpus and a corresponding KG, We denote a token sequence in the corpus as $\{w_1,\ldots,w_n\}$, where $n$ is the length of the token sequence. Meanwhile, the entity span sequence aligning to the given tokens is $\{e_1,\ldots,e_m\}$, where $m \leq n$ is the number of entities contained in the token sequence. Each entity span is described by its start
$\mathrm{si}$ and end $\mathrm{ei})$ %index
indices
in the token sequence $e_i=(\mathrm{si}, \mathrm{ei})$.
% \mathrm{start}, \mathrm{end})$.
%Besides, 
Finally, let 
$\{r_{ij}\}$ %is 
be the relation between entities $e_i$ and $e_j$ when the entities 
% have relation 
are related in the KG, where 
%$i, j \in {1,\ldots,m}$.
$1 \leq i < j \leq m$. \our will output the embeddings $\mathbf{w}$, $\mathbf{e}$, $\mathbf{r}$ to represent tokens, entities and relationships respectively.
\subsection{Model Architecture}
\label{sec:model_arch}
%As shown in Figure \todo{framework of model}, the 
%whole
As shown in Figure~\ref{fig:overview}, \our consists of three components:
\begin{enumerate}
    \item a textual encoder (\texttt{TextEnd}) responsible for capturing lexical and syntactic information from the input tokens;
    \item a span encoder (\texttt{SpanEnd}) that learns to generate the representation of contiguous tokens (spans) in the text as an entity; and
    \item a span pair encoder (\texttt{PairEnd}) that learns to generate the representation of span pairs capturing relation information between spans.
\end{enumerate}

% Based on the three components, %three-level
Each component (i.e., token, span, and span pair encoders) generate representations that accommodate a different information/knowledge type (i.e., word, entity, relation) in the text corpus and KG. 
% We next provide an overview of the end-to-end process before describing the span and span pair encoders in more detail:\\

{\bf Textual encoder.} Given a token sequence $\{w_1,\ldots,w_n\}$ and its corresponding entity spans $\{e_1,\ldots,e_m\}$, 
the textual encoder first sums the token embedding, segment embedding, and positional embedding for each token to compute its input embedding, and then computes lexical and syntactic features $\{\mathbf{w}_1,\ldots,\mathbf{w}_n\}$ by a multi-head self-attention mechanism by: % as follows:
\begin{equation}
    \{\mathbf{w}_1,\ldots,\mathbf{w}_n\} = \texttt{TextEnd}(\{w_1,\ldots,w_n\})
\end{equation}
where $\mathrm{TextEnd}$ is a multi-layer bidirectional Transformer, identical to its implementation in BERT~\cite{Devlin2019BERTPO}.\\

{\bf Span encoder.} Different from previous works that used separate embeddings for each 
%entities 
entity
in the KG, span representations are computed from token-level features $\{\mathbf{w}_1,\ldots,\mathbf{w}_n\}$ by \texttt{SpanEnd} for all entity spans $\{e_1,\ldots,e_m\}$ to 
% get 
obtain their entity representations $\{\mathbf{e}_1,\ldots, \mathbf{e}_m\}$:
\begin{equation}
\begin{split}
     \{\mathbf{e}_1,\ldots, \mathbf{e}_m\} =& \texttt{SpanEnd}(\{\mathbf{w}_1,\ldots,\mathbf{w}_n\},\\ &\{e_1,\ldots,e_m\})
\end{split}
\end{equation}\\
Details of the span encoder \texttt{SpanEnd}
will be described
in Section~\ref{sec:span}. 

{\bf Span pair encoder.} To further capture the relations between entities in the KG, \texttt{PairEnd} computes relation representations between any two pairs of spans:
\begin{equation}
    \mathbf{r}_{ij} = \texttt{PairEnd}(\{\mathbf{e}_i,\mathbf{e}_j\}).
\end{equation}
$\mathbf{r}_{ij}$ is the relation representation of entities $e_i$ and $e_j$.\\
% 
%More d
Details of % the span encoder \texttt{SpanEnd} and 
pair encoder \texttt{PairEnd} will be described in
% introduced in Section~\ref{sec:span} and 
Section~\ref{sec:pair}.  % respectively.

% We next focus on the span and span pair encoders.

\subsection{Span Encoder}
\label{sec:span}
As explained above, the goal of \texttt{SpanEnd} is to compute the span representations $\{\mathbf{e}_1,\ldots, \mathbf{e}_m\}$ from the token representations $\{\mathbf{w}_1,\ldots,\mathbf{w}_n\}$ and the entity spans $\{e_1,\ldots,e_m\}$. In this work, we explored three different methods % for getting information about the spans in the token sequence.
to learn the representations of entity spans.

\textbf{Boundary points.}  Span representations with boundary information 
%is 
were
first applied to question answering~\cite{Lee2016LearningRS}. We referred to this method  
%choice 
as the \texttt{EndPoint} representation. In this method, the textual encoder outputs are concatenated at the endpoints of a span to jointly encode its inside and outside information. To distinguish different spans sharing the same endpoints %(e.g., spans \texttt{New York City} and \texttt{New City}), 
%(e.g., spans \texttt{New York State} and \texttt{New Mexico State}),
(e.g., spans \texttt{``deep unsupervised learning''} and \texttt{``deep learning''}),
the entity width embedding $\mathbf{E_w}=\{e_{w}^{1},\ldots,e_{w}^{l}\}$ is concatenated with the endpoint representation, where $\mathbf{E_w}\in \mathbb{R}^{l\times d}$, $l$ is the max length of the spans, $d$ is the dimension of the embeddings,
% embedding dimension:
and where $\mathrm{si}$ and $\mathrm{ei}$ are the start
and end positions of span $e_i$.
\begin{equation}
    \mathbf{e}_i^{\mathrm{endpoint}} = \left[\mathbf{w}_{si}, \mathbf{w}_{ei}, e_w^{(ei-si+1)}\right]
\end{equation}

\textbf{Self-attention.} \citet{Lee2017EndtoendNC} 
described a method
to learn a task-specific notion of a span head using an attention mechanism over words in each span. Given token representations $\{\mathbf{w}_1,\ldots,\mathbf{w}_n\}$ and a span $e_i=(\mathrm{start}, \mathrm{end})$, a self-attentive span representation is shown as 
%following:
follows:
\begin{align}
    \alpha_j &= \mathrm{FFNN}(\mathbf{w}_j) \\
    a_{i,j} &= \frac{\mathrm{exp}(\alpha_j)}{\sum_{k = \mathrm{start}}^\mathrm{end} \mathrm{exp}(\alpha_k)} \\
    \mathbf{e}_i^{\mathrm{selfattn}} &= \sum_{j= \mathrm{start}}^\mathrm{end} a_{i,j}\cdot \mathbf{w}_j
\end{align}
where $\mathbf{e}_i^{\mathrm{selfattn}}$ is a weighted sum of 
the
token representations 
of tokens
in an entity span $e_j$. The weights $a_{i,j}$ are automatically learned and $\mathrm{FFNN}(\cdot)$ is a feed-forward neural network.

\textbf{Max Pooling.} The span representation $\mathbf{e}_i^{\mathrm{pooling}}$ is obtained by maxpooling the token representations corresponding to each span. To be specific, $\mathbf{e}_i^{\mathrm{pooling}}=\mathrm{MaxPool}([\mathbf{w}_{\mathrm{si}},\ldots,\mathbf{w}_{\mathrm{ei}}])$,
where $\mathrm{si}$ and $\mathrm{ei}$ are the 
%begin 
start
and end positions of span $e_i$.

In our experiments the concatenation of $\mathbf{e}_i^{\mathrm{endpoint}}$ and $\mathbf{e}_i^{\mathrm{selfattn}}$ yielded the best results, which is why we adopt this as \our's final span representation: $\mathbf{e}_i=[\mathbf{e}_i^{\mathrm{endpoint}}, \mathbf{e}_i^{\mathrm{selfattn}}]$. Results of the comparison between the different span representation methods can be found in % the \emph{Ablation Study} 
Section~\ref{sec:ablation}. 

%In this paper, we adopt the concatenation of $\mathbf{e}_i^{\mathrm{endpoint}}$ and $\mathbf{e}_i^{\mathrm{selfattn}}$ as the final representation of span $\mathbf{e}_i=[\mathbf{e}_i^{\mathrm{endpoint}}, \mathbf{e}_i^{\mathrm{selfattn}}]$. We will also discuss the best span representation among the above methods and their combinations in Section~\ref{sec:ablation}.

\subsection{Pair Encoder}
\label{sec:pair}
The pair encoder is responsible for generating a representation of the relation given two spans in a sentence. Relations between two spans are usually determined by both the entity types of the spans and their context in a sentence.  
% In 
% Previous, 
% DyGIE~\cite{Luan2019AGF}
% % the authors apply 
% also
% modeled a relationship 
% % interactions 
% between spans based on their context
% but used graphs.
% and contain more contextual information. 

In this paper, 
self-attention is used
to encode span representations efficiently with the contextual information of the spans. 
Specifically, span representations are sent to self-attention layers and then use concatenation of two contextual span representations to construct the final span pair representation:
\begin{align}
    \{\mathbf{\Tilde{e}}_{1},\ldots,\mathbf{\Tilde{e}}_{m}\} &= \mathrm{SelfAttn}(\{\mathbf{e}_{1},\ldots,\mathbf{e}_{m}\}) \\
    \mathbf{r}_{ij} &= \mathrm{FFNN}([\mathbf{\Tilde{e}}_{i}, \mathbf{\Tilde{e}}_{j}]), % \\
    % 1 & \leq i < j \leq m, \nonumber
\end{align}
where $\mathrm{SelfAttn}(\cdot)$ is a multi-head self-attention mechanism and $\mathrm{FFNN}(\cdot)$ is a feed forward neural network.

\subsection{Pre-training Task}
\label{sec:pre_training}
To hierarchically learn representations from tokens, entities and relations at the same time. We propose a three-level pre-training task for our model:
(1) token level,
(2) entity level, and
(3) relation level.

At the \emph{token level},
similar to BERT, \our adopts the masked language model (MLM) but has different masking strategies on tokens. The intuition is that the model can learn to infer both masked tokens and masked entities in the input sentence based on their contexts and other entities.
To this end, two masking strategies are performed:
\begin{enumerate}
    \item token masking: randomly mask $10\%$
of tokens in sentences as other masked language models (e.g.,~BERT);
    \item entity masking: randomly mask $20\%$
of entities~\footnote{All tokens in the entity are masked if an entity is selected to be masked.} in sentences.
\end{enumerate}

Following previous works, $10\%$ of masked tokens are replaced with randomly selected tokens and keep $10\%$ of tokens unchanged. The objective is the log-likelihood that maximizes the probability of masked tokens and is denoted by:
\begin{equation}
    \mathcal{L}_{\mathrm{MLM}} = -\sum_{w^*\in \mathcal{M}}\mathrm{log}P(w^*|\mathbf{w}_i)
\end{equation}
where $w^*$ is the masked token introduced in our two masking strategies and the masked token set is denoted by % represented as 
$\mathcal{M}$; $\mathbf{w}_i$ is the token representation from the token level of \our.

At the \emph{entity level}, entity prediction is based on the span encoder's entity representation $\mathbf{e}_i$. Because of the large number and unbalanced distribution of entities, entity-level loss is computed by the adaptive softmax~\cite{Grave2017EfficientSA} with log-likelihood for pre-training,
\begin{equation}
\mathcal{L}_{\mathrm{ENT}} = -\sum_{e^*\in \mathcal{D}}\mathrm{log}P(e^*|\mathbf{e}_i)
\end{equation}
where $e^*$ are gold entities in training documents $\mathcal{D}$. To endow the model with the ability to recognize entities from spans that do not express any entities. For each training instance, spans are randomly sampled as negatives which have the same number as entities in a sentence. Specifically, all spans up to the max length $8$ are enumerated in our pre-training. Because most spans do not express any entities, we assumed that the random sampling will not sample any entities and used the sampled spans as negatives.

At the \emph{relation level}, relationships are predicted between entity $i$ and entity $j$ from the entity pair representation $\mathbf{r}_{ij}$. Similar to tokens and entities, the log-likelihood is calculated to maximize the probability of the relation between two entities. The loss for relation $\mathcal{L}_{\mathrm{REL}}$ is calculated by:
\begin{equation}
\mathcal{L}_{\mathrm{REL}} = -\sum_{r^*\in \mathcal{D}}\mathrm{log}P(r^*|\mathbf{r}_i)
\end{equation}
where $r^*$ are ground-truth relations in the training documents $\mathcal{D}$.

The entire network is trained to convergence by minimizing the summation of the three losses.
\begin{equation}
    \mathcal{L} = \mathcal{L}_{\mathrm{MLM}} + \mathcal{L}_{\mathrm{ENT}} + \mathcal{L}_{\mathrm{REL}}
\end{equation}

%% file: 4_experiment.tex
\subsection{Pre-training Dataset}
We used English Wikipedia as our pre-training corpus and aligned the 
text to Wikidata. Text with hyperlinks will be linked to an entity in Wikidata and two entities will be assigned with a relation in one sentence if they have a property in Wikidata. After discarding sentences without any relations, we used a subset of the whole dataset for pre-training. The subset corpus includes 31,263,279 sentences, 10,598,458 entities and 1,002
%$1002$ 
relations. For efficient pre-training, we only included the most frequent $1$M entities in the corpus. Due to the unbalanced distribution of entities, the $1$M entities cover $80\%$ of phrases with hyperlinks in Wikipedia.

\subsection{Parameter Settings and Training Details}
The backbone model of our textual encoder is $\text{RoBERTa}_{\text{large}}$. We implemented our method using Huggingface's Pytorch transformers~\footnote{https://huggingface.co/transformers/}. The total amount of parameters of RoBERTa is about $355$M. The span module and span pair module have $16$M and $5$M parameters respectively. We can see that our knowledge modules are much smaller than the language module.  In the span encoder, the max length of spans is set to $8$ and spans longer than this value are truncated. Two-layer self-attention with $8$ heads is applied to obtain contextual span representations in the pair encoder. The hidden size of the span and pair encoder is 1,024 (same for the textual encoder). We pre-trained our model on the Wiki corpus for three epochs. To accelerate the training process, the max sequence length is reduced from $512$ to $256$ as the computation of self-attention is quadratic in the length. The batch size for pre-training is $96$. We set the learning rate as 5e-5 and optimize our model with Adam. We pre-trained the model with three 32GB NVIDIA Tesla V100 GPUs for $23$ days.

For all information extraction tasks, we used the Adam optimizer and select the best learning rate from \{1e-5, 2e-5, 5e-5\}, the best warm-up steps from \{$0, 300, 1000$\} and the best weight decay from \{$0.01$, 1e-5\}. We used a single GPU for all fine-tuning on downstream information extraction tasks. Linear learning rate decay strategy was adopted and gradient was clipped to $5$ for all experiments. The dropout ratio was $0.1$ for language model and $0.3$ for task-specific layers. The batch size was $16$ for joint learning entities and relationships, entity typing, and $32$ for relation extraction task. We adopted early stop by using the best model on a development set before tested the model on a test set. 

\subsection{Clustering with Pre-trained Embeddings}
To assess the quality of the representations of entities and relationships learned by~\our,
% performance of knowledge representations from pre-trained language models, in this section, 
we conducted entity embedding clustering and relation embedding clustering and compared to a set of competitive language models as our baselines.
% to show the ability to representing knowledge.

\subsubsection{Entity Clustering}
\label{sec:entity_cls}
Entity clustering is conducted on BC5CDR~\cite{Li2016BioCreativeVC}, which is the BioCreative V Chemical and Disease Recognition task corpus. It contains 18,307 sentences from PubMed articles, with 15,953 chemical and 13,318 disease entities.

We compared~\our to the following baselines:
\begin{itemize}
    \item \textbf{GloVe}~\cite{Pen2014Glove}. Pre-trained word embeddings on 6B tokens and the dimension is $300$. We used averaged word embeddings as its entity representations.
    \item \textbf{BERT}~\cite{Devlin2019BERTPO}. Option 1: averaging token representations in an entity span (BERT-Ave.); 2: substituting entities with \texttt{[MASK]} tokens, and use \texttt{[MASK]} representations generated as the entity embeddings (BERT-MASK); or 3: concatenating representations of the first and the last tokens as the entity embeddings (BERT-End.).
    \item \textbf{LUKE}~\cite{Yamada2020LUKEDC}. The contextual entity representations from LUKE are used. % as entity features in our clustering.
    \item \textbf{ERICA}~\cite{Qin2021ERICAIE}. % Pre-trained model that 
    Applying the mean-pooling operation over their representations generated for consecutive tokens to obtain local entity representations.
\end{itemize}
For SPOT, the outputs from span module represented entities. K-Means was applied to create the clusters for each entity. We followed previous clustering work~\cite{Xu2017SelfTaughtCN} and adopted Accuracy (ACC), Normalized Mutual Information (NMI), and Adjusted Rand Index (ARI)~\cite{Steinley2004PropertiesOT} to evaluate the quality of the clusters from different models. % approaches.

\begin{table}[t]
\vspace{-2mm}
\caption{Entity clustering results on BC5CDR.}
\vspace{-2mm}
\begin{tabular}{lccc}
\toprule
\textbf{Metrics} & \textbf{ACC}   & \textbf{NMI}    & \textbf{ARI}   \\ \midrule
GloVe     & 0.587 & 0.026 & 0.030 \\
BERT-Ave. & 0.857 & 0.489  & 0.510 \\
BERT-MASK & 0.551 & 0.000  & 0.002 \\
BERT-End. & 0.552 & 0.000  & 0.003 \\
LUKE      & 0.794 & 0.411  & 0.346 \\
ERICA     & 0.923 & 0.628  & 0.715 \\
SPOT      & \textbf{0.928} & \textbf{0.645}  & \textbf{0.731} \\ \bottomrule
\end{tabular}
\vspace{-4mm}
\label{tab:entity_cls}
\end{table}

% We report
Table~\ref{tab:entity_cls} shows the evaluation results of the entity clustering. % Overall, 
In all metrics, 
\our achieved the best results compared to all baselines. Without any label, \our was able to distinguish disease entities and chemical entities with an accuracy of $0.928$. Although ERICA achieved a close accuracy to \our, % there are still 
clear margins on NMI and ARI % are 
% which 
indicate that entities of 
% representations in 
the same class might
% will be 
were more concentrated with \our than ERICA. For BERT, average pooling was the best method to represent entities from token representations.
\our outperformed LUKE though LUKE was designed specifically for entity representations with a large number of parameters to encode every entity.
Averaging GloVe, BERT-Ave. and BERT-End. cannot provide effective entity representations. From the results, we can see that our pre-trained span encoder can output high-quality entity representations from token representations. 

\subsubsection{Relation Clustering}
Relation clustering was performed on NYT24~\cite{Zeng2018ExtractingRF}, which contains 66,196 sentences and $24$ relationships between entities distantly labeled by a knowledge base. In this experiment, we randomly sampled $4$ relationships from the whole dataset and then clustered relationships into $4$ groups. The sampling and clustering was repeated $5$ times to reduce the bias of relationship selection.

We compared~\our to the following baselines:
\begin{itemize}
    \item \textbf{GloVe}~\cite{Pen2014Glove}. The entity representations introduced in Section~\ref{sec:entity_cls} were used as relation representations.
    \item \textbf{BERT}~\cite{Devlin2019BERTPO}. Two options: 1. Obtaining relation representations by concatenating entity representations in BERT-Ave.; 2. concatenating entity representations from BERT-End.
    \item \textbf{RoBERTa}~\cite{Liu2019RoBERTaAR}. Using the same method as BERT to obtain relation representations but loading the pre-trained parameters from RoBERTa.
    \item \textbf{LUKE}~\cite{Yamada2020LUKEDC}. Concatenating entity representations as relation representations.
    \item \textbf{ERICA}~\cite{Qin2021ERICAIE}. The same relation representations as introduced by the authors were adopted.
\end{itemize}
For SPOT, the outputs from the span pair module represent relationships. The clustering method and evaluation metrics were the same as Section~\ref{sec:entity_cls}.

\begin{table}[t]
\vspace{-2mm}
\caption{Relation clustering results on NYT24. The averaged results are from $5$ independently repeated experiments.}
\vspace{-2mm}
\begin{tabular}{lccc}
\toprule
\textbf{Metrics} & \textbf{ACC}   & \textbf{NMI}    & \textbf{ARI}   \\ \midrule
GloVe     & 0.493 & 0.282 & 0.190 \\
BERT-Ave. & 0.501 & 0.230  & 0.134 \\
BERT-End. & 0.504 & 0.228  & 0.113 \\
RoBERTa-Ave. & 0.488 & 0.240  & 0.185\\
RoBERTa-End. & 0.493 & 0.122  & 0.089 \\
LUKE      & 0.557 & 0.388 & 0.274 \\
ERICA     & 0.711 & 0.525  & 0.449 \\
SPOT      & \textbf{0.756} & \textbf{0.533}  & \textbf{0.453} \\ \bottomrule
\end{tabular}
\vspace{-2mm}
\label{tab:relation_cls}
\end{table}

The results of relation clustering are shown in Table~\ref{tab:relation_cls}. Overall, \our achieved the best results compared to all baselines. We can see that knowledge-enhanced models % methods 
(e.g.,~LUKE, ERICA and~\our) largely outperform general language models (e.g.,~BERT, RoBERTa and GloVe). Because ERICA and~\our incorporated both entity and relation knowledge during pre-training, these two models performed significantly better than LUKE on relation clustering.

From the entity and relation clustering, we can see that~\our can effectively represent entities and relationships by the span modules.

% \subsection{Training Details for Fine-tuning}
% For all information extraction tasks, we used the Adam optimizer and select the best learning rate from \{1e-5, 2e-5, 5e-5\}, the best warmup steps from \{$0, 300, 1000$\} and the best weight decay from \{$0.01$, 1e-5\}. We used a single GPU for all fine-tuning on downstream information extraction tasks. Linear learning rate decay strategy is adopted and gradient is clipped to $5$ for all experiments. The dropout ratio is $0.1$ for language model and $0.3$ for task-specific layers. The batch size is $16$ for joint learning entities and relationships, entity typing, and $32$ for relation extraction task. We adopted early stop by using the best model on development set and test on test set. 

\subsection{Joint Entity and Relation Extraction}
Given a sentence, joint entity and relation extraction methods both extract entities in the sentence and predict relations between these entities. A span and span pair framework which is consistent with our pre-training framework was applied for this task. All models were fine-tuned and evaluated on WebNLG~\cite{Gardent2017CreatingTC} which contains $216$ relation types. The maximum length of spans we considered was $5$. We compared our model with the following models:

\begin{itemize}
\item \textbf{TPLinker}~\cite{Wang2020TPLinkerSJ}, which is a % typical 
benchmark method used in joint entity and relation extraction;
\item \textbf{TDEER}~\cite{Li2021TDEERAE}, which is considered as the state-of-the-art method in this task and the backbone is BERT;
\item \textbf{BERT}~\cite{Devlin2019BERTPO} and \textbf{RoBERTa}~\cite{Liu2019RoBERTaAR}, which are widely used as text encoders for various tasks;
\item \textbf{SpanBERT}~\cite{Joshi2020SpanBERTIP} which masks and predicts contiguous random spans instead of random tokens;
\item \textbf{CorefBERT}~\cite{Ye2020CoreferentialRL} is a pre-training method that incorporates the coreferential relations in context;
\item \textbf{ERICA}~\cite{Qin2021ERICAIE} improves entity and relation understanding by contrastive learning.
\end{itemize}
% In 
Among these baselines, \textbf{TPLinker} and \textbf{TDEER} are task-specific methods designed for joint entity and relation extraction; \textbf{BERT} and \textbf{RoBERTa} are language models for general purposes; \textbf{SpanBERT}, \textbf{CorefBERT} and \textbf{ERICA} are knowledge-enhanced language models.
For a fair comparison with other language models, we did not load pre-trained parameters of span and span pair encoder in~\our but trained them as a part of the fine-tuning.

\begin{table}[t]
\vspace{-2mm}
\caption{Results on joint entity and relation extraction (WebNLG). We reported precision, recall and micro F1.}
\vspace{-2mm}
\begin{tabular}{lccc}
\toprule
\textbf{Metrics}         & \textbf{Precision} & \textbf{Recall} & \textbf{F1}   \\ \midrule
TPLinker  & 91.8      & 92.0   & 91.9 \\
TDEER     & \textbf{93.8}      & 92.4   & 93.1 \\ \midrule
BERT      & 91.3      & 92.5   & 91.8 \\
RoBERTa   & 91.3      & 92.5   & 91.1 \\
SpanBERT  & 92.1      & 91.9   & 92.0 \\
CorefBERT & 91.8      & 92.6   & 92.2 \\
ERICA     & 91.6      & 92.6   & 92.1 \\ \midrule
SPOT      & 93.4      & \textbf{93.1}   & \textbf{93.3} \\ \bottomrule
\end{tabular}
\vspace{-2mm}
\label{tab:joint}
\end{table}

Results in Table~\ref{tab:joint} show that~\our achieved the highest Recall and F1 compared to task-specific models and other language models. Specifically, task-specific models performed better than other language models because their designed modules for this task. Knowledge-enhanced language models outperformed general language models, which indicates the effectiveness of incorporating knowledge into language models for joint entity and relation extraction. \our significantly outperformed all language models, which demonstrates that~\our % better understand 
represented the knowledge better with our span and span pair modules.

\subsection{Entity Typing}
Entity typing aims %is aiming 
at predicting the type of an entity given a sentence and its position. All models were trained and evaluated on the dataset FIGER~\cite{Ling2015DesignCF} which contains $113$ entity types, $2$ million and $10,000$ distantly labeled sentences for training and validation; and $563$ human labeled sentences as the test set.
We compared our model with \textbf{BERT}, \textbf{RoBERTa}, \textbf{SpanBERT}, \textbf{CorefBERT}, \textbf{ERICA} and the following models:

\begin{itemize}
\item \textbf{ERNIE} incorporates knowledge graph information into BERT to enhance entity representations;
\item \textbf{LUKE} treats words and entities in a given as independent tokens and outputs contextualized representations of them;
\item \textbf{WKLM} employs a zero-shot fact completion task to improve pre-trained language models by involving knowledge.
\end{itemize}
Following \citet{Zhang2019ERNIEEL}, two special tokens \texttt{[ENT]} are inserted into sentences to highlight entity mentions for language models such as~\textbf{BERT}~\footnote{For LUKE, we used the their contextual entity representations to predict the types; results of ERICA, ERNIE and WKLM were from their papers; other baselines used inserted special tokens to represent entities.}. 

\begin{table}[t]
\vspace{-2mm}
\caption{Results on entity typing (FIGER). We reported precision, recall and micro F1 on the test set.}
\vspace{-2mm}
\begin{tabular}{lccc}
\toprule
\textbf{Metrics}         & \textbf{Precision} & \textbf{Recall} & \textbf{F1}   \\ \midrule
BERT      & 66.4      & 88.5   & 75.8 \\
RoBERTa   & 65.1      & 88.1   & 74.9 \\
SpanBERT  & 66.4      & 79.9   & 72.5 \\
ERNIE     & -         & -      & 73.4 \\
LUKE      & \textbf{69.9}      & 89.0   & \textbf{78.3} \\
WKLM      & -      & -        & 77 \\
CorefBERT & 62.4      & 82.2   & 72.2 \\
ERICA     & -      & -   & 77.0 \\ \midrule
SPOT      & 68.5      & \textbf{89.2}   & 77.5 \\ \bottomrule
\end{tabular}
\vspace{-3mm}
\label{tab:typing}
\end{table}

From the results listed in Table~\ref{tab:typing} we observe that, overall, knowledge-enhances models (e.g.,~LUKE, WKLM, ERICA) achieved significant improvements compared to general language models (e.g.,~BERT, RoBERTa, SpanBERT). LUKE achieved the best precision and F1 and \our achieved the best recall. Compared to our backbone RoBERTa,  \our improved RoBERTa by $2.6$ (F1) which indicates the effectiveness of our span modules for knowledge incorporation.

\subsection{Relation Extraction}
Relation extraction aims at determining the correct relation between two entities in a given sentence. We evaluated all models on SemEval2010~~\cite{hendrickx-etal-2010-semeval} which contains $18$ relation types, 8,000 sentences for training and 2,717 sentences for test. Macro F1 was used for SemEval2010 as the official evaluation. We compared~\our with \textbf{BERT}, \textbf{RoBERTa}, \textbf{SpanBERT}, \textbf{CorefBERT}, \textbf{LUKE}, \textbf{ERICA} and three following models:
\begin{itemize}
    \item \textbf{KnowBERT}~\cite{Peters2019KnowledgeEC} outputs contextual word embeddings enhanced by entity representations from knowledge graphs.
    \item \textbf{MTB}~\cite{Soares2019MatchingTB} is a pre-trained model designed for relation extraction by distinguishing if the two sentences express the same relationship.
    \item \textbf{CP}~\cite{Peng2020LearningFC}, a contrastive learning method that trains models on distantly labeled datasets.
\end{itemize}
To evaluate our model on this task, following~\citet{Peters2019KnowledgeEC}, different tokens \texttt{[HD]} and \texttt{[TL]} were inserted for head entities and tail entities respectively and the contextual word representations for \texttt{[HD]} and \texttt{[TL]} were concatenated to predict the relationship.

\begin{table}[t]
\vspace{-2mm}
\caption{Results on relation extraction (SemEval2010). We reported precision, recall and macro F1 on the test set.}
\vspace{-2mm}
\begin{tabular}{lccc}
\toprule
\textbf{Metrics}         & \textbf{Precision} & \textbf{Recall} & \textbf{Macro F1}   \\ \midrule
BERT      & 88.6      & 90.4   & 89.4 \\
RoBERTa   & 88.4      & 89.0   & 88.7 \\
SpanBERT  & 87.9      & 89.7   & 88.8 \\
KnowBERT     & 89.1         & 89.1      & 89.1 \\
MTB       & 88.1          & 90.1       & 89.2 \\
CP        & 88.6          & 90.4       & 89.5 \\
CorefBERT & 89.2         & 89.2     & 89.2 \\
LUKE      & 89.3      & 91.3   & 90.3 \\
ERICA     & 89.6      & 89.0   & 89.2 \\ \midrule
SPOT      & \textbf{89.9}     & \textbf{91.4}   & \textbf{90.6} \\ \bottomrule
\end{tabular}
\vspace{-2mm}
\label{tab:relation}
\end{table}

Results in Table~\ref{tab:relation} show that~\our outperformed all baselines and LUKE achieved similar results as~\our. We can still see some improvements by incorporating external knowledge into language models because the F1 scores of all knowledge-enhanced models were larger than $89$.

\label{sec:repr}
\begin{figure*}[t]
\scalebox{0.9}{
     \centering
     \begin{subfigure}{0.245\textwidth}
         \centering
         \includegraphics[width=\textwidth]{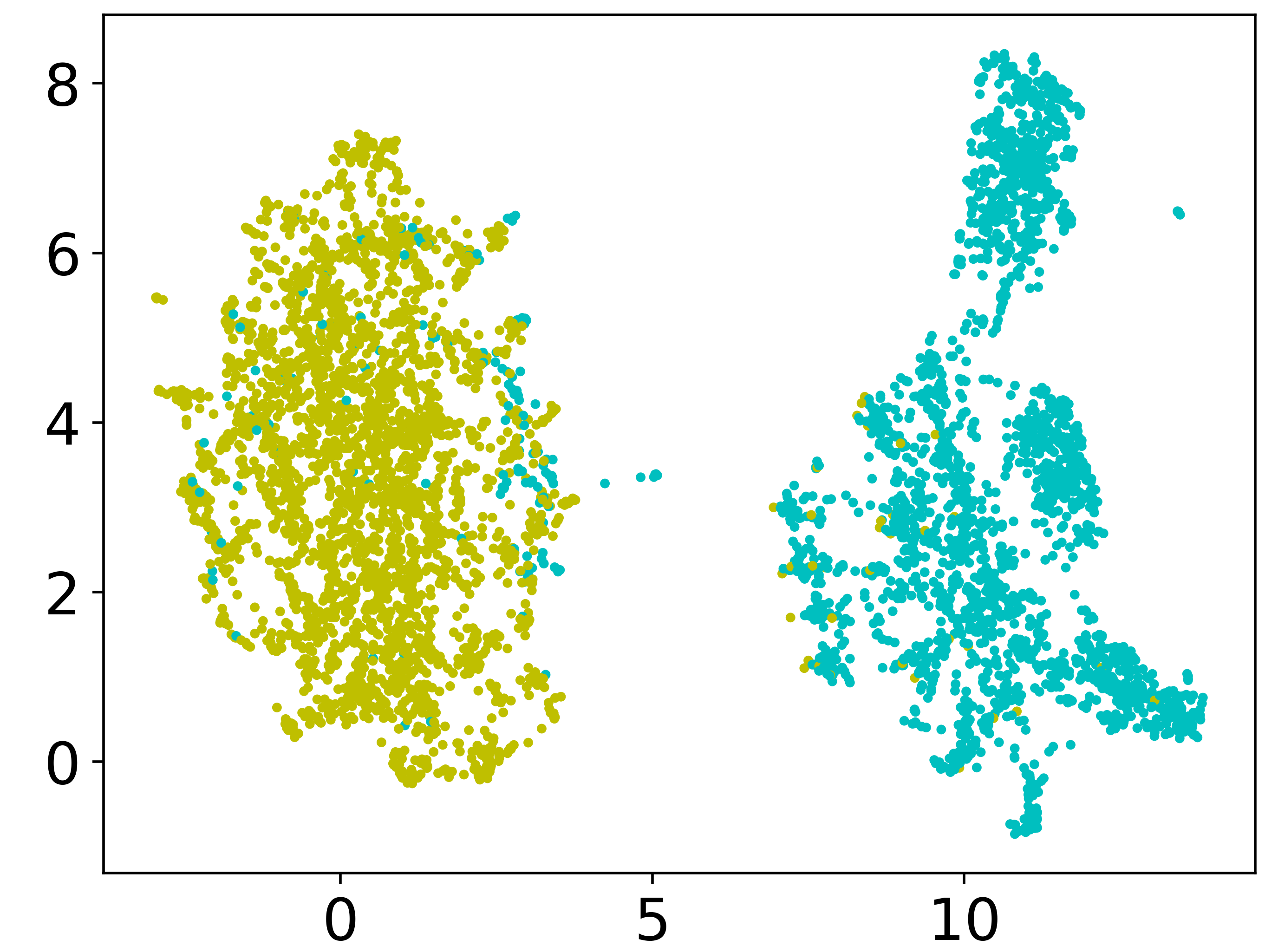}
         \caption{\textbf{SPOT}}
         \label{fig:bc5cdr}
     \end{subfigure}
     \hfill
     \begin{subfigure}{0.245\textwidth}
         \centering
         \includegraphics[width=\textwidth]{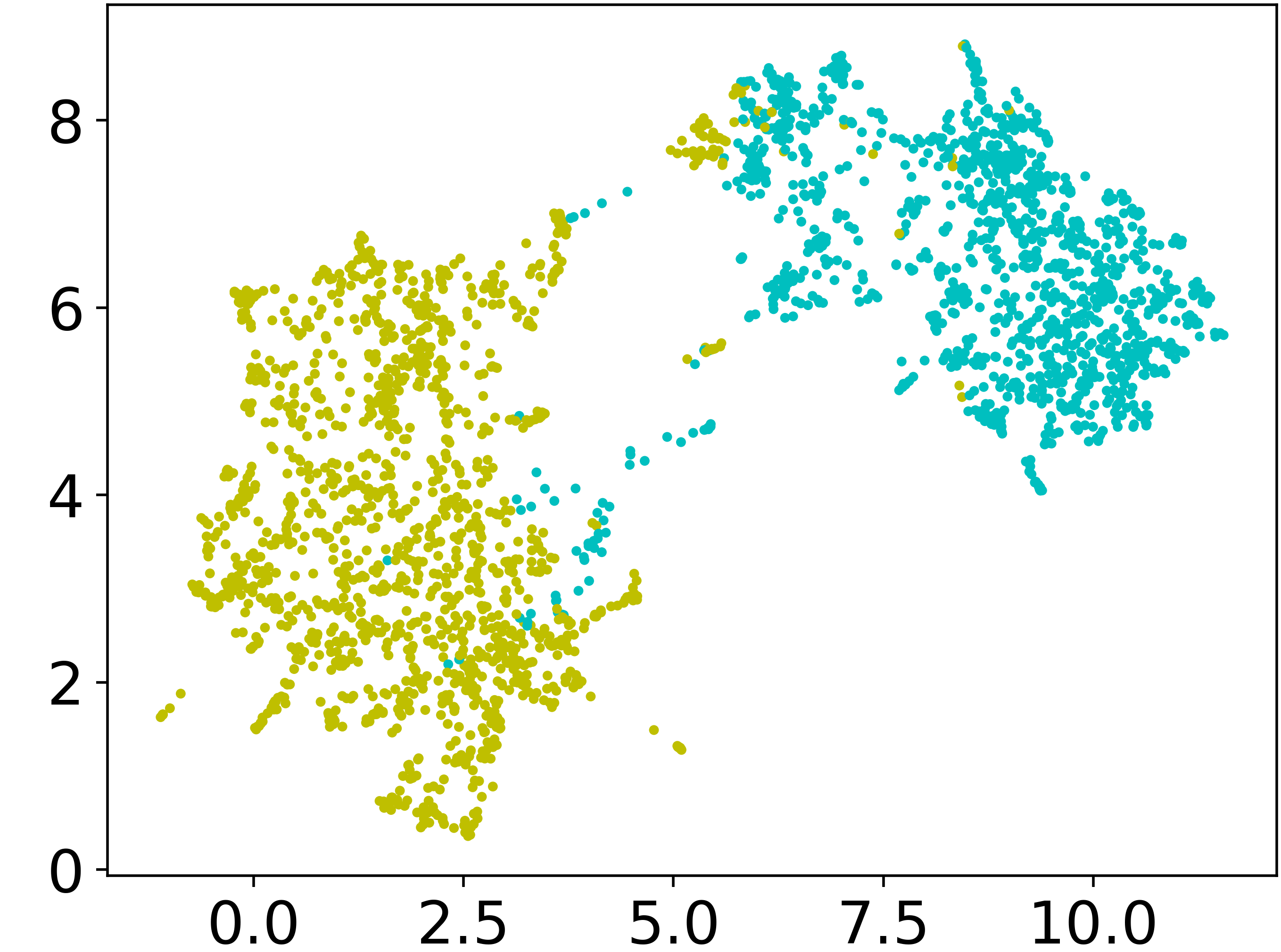}
         \caption{LUKE}
         \label{fig:bc5cdr_luke}
     \end{subfigure}
     \hfill
     \begin{subfigure}{0.245\textwidth}
         \centering
         \includegraphics[width=\textwidth]{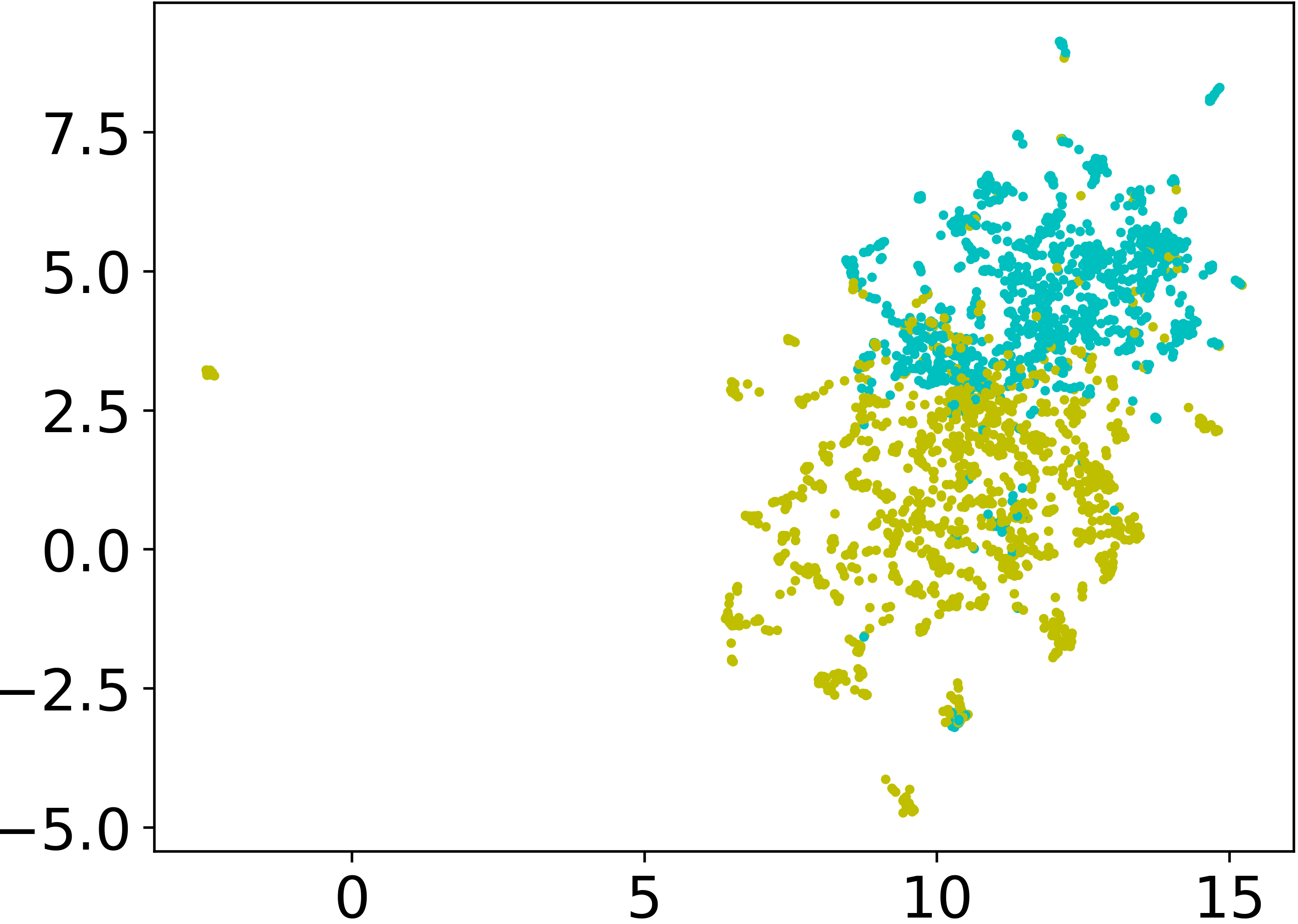}
         \caption{RoBERTa-endpoint}
         \label{fig:bc5cdr_endpoint}
     \end{subfigure}
     \hfill
     \begin{subfigure}{0.245\textwidth}
         \centering
         \includegraphics[width=\textwidth]{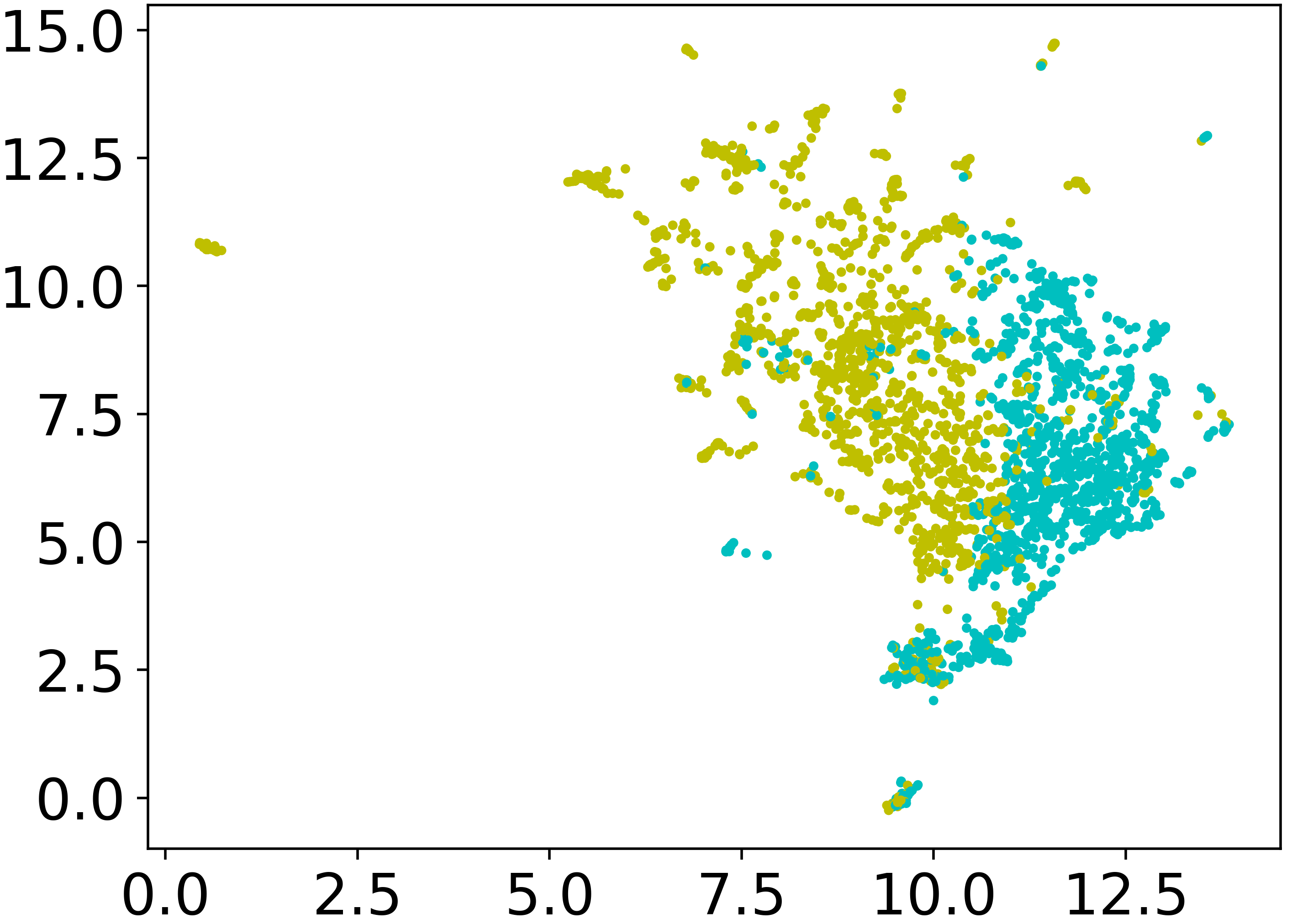}
         \caption{RoBERTa-maxpool}
         \label{fig:bc5cdr_maxpooling}
     \end{subfigure}
     }
     \\
     \centering
     \scalebox{0.9}{
     \begin{subfigure}{0.245\textwidth}
         \centering
         \includegraphics[width=\textwidth]{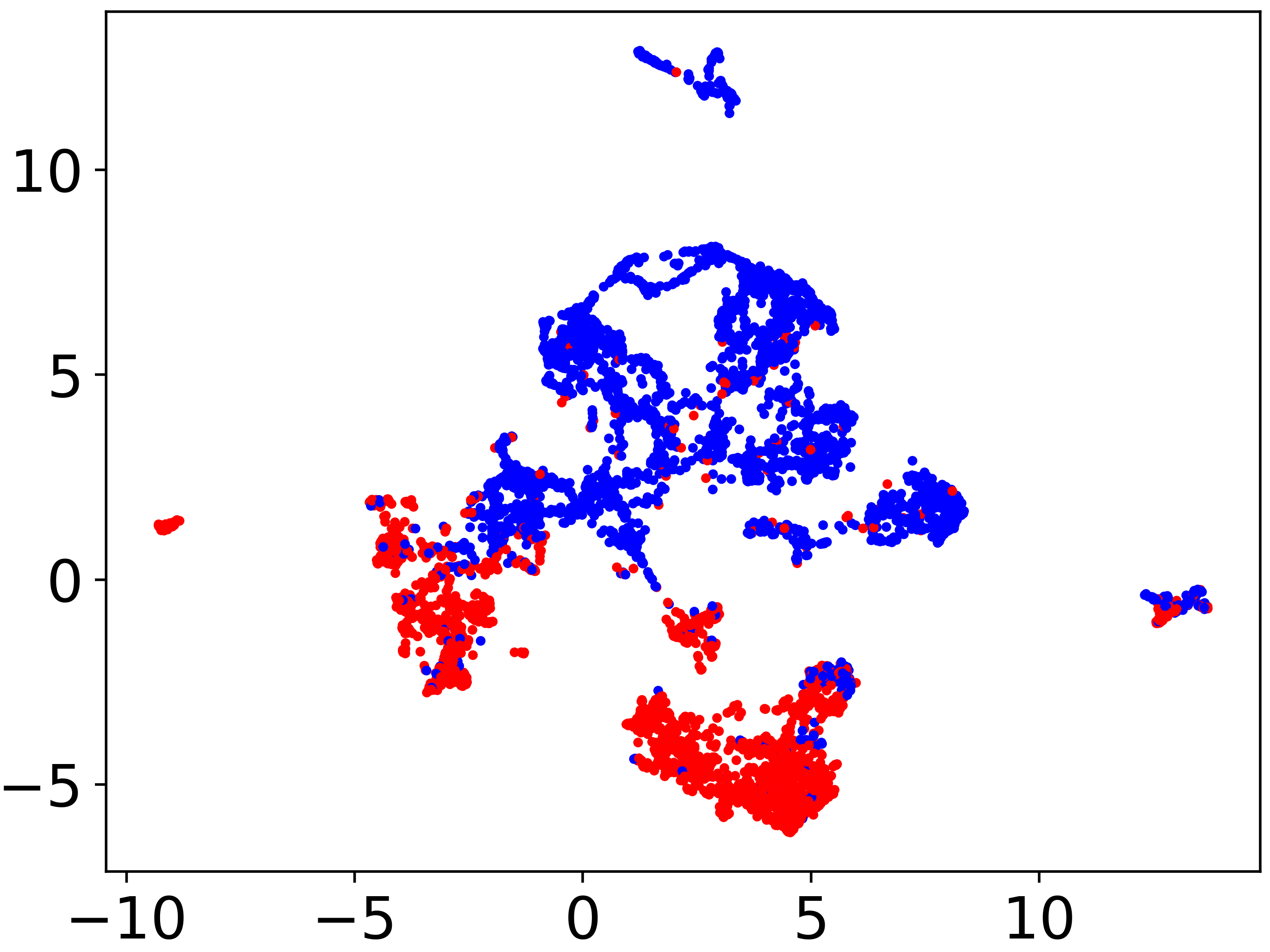}
         \caption{\textbf{SPOT}}
         \label{fig:conll}
     \end{subfigure}
     \hfill
     \begin{subfigure}{0.245\textwidth}
         \centering
         \includegraphics[width=\textwidth]{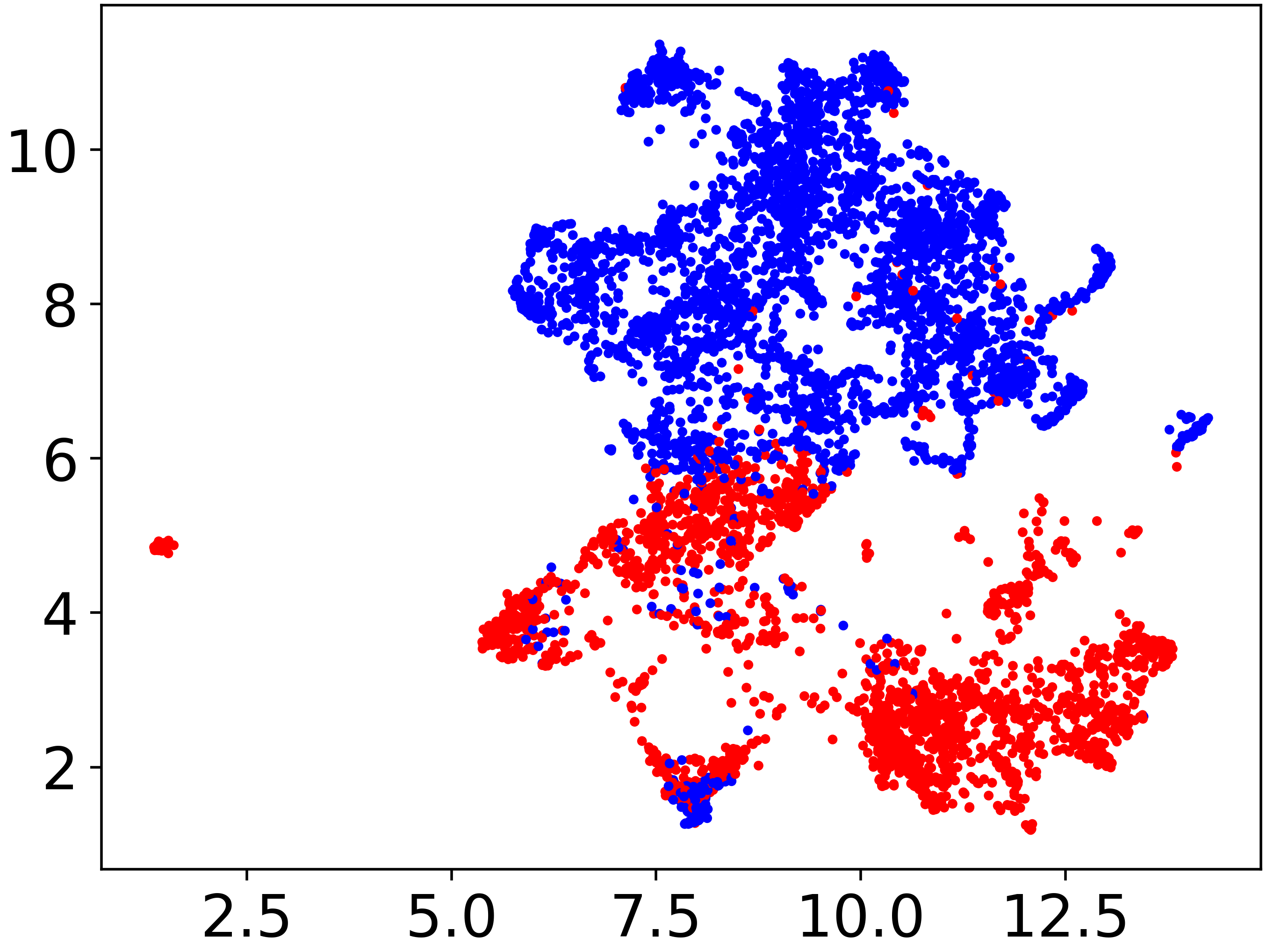}
         \caption{LUKE}
         \label{fig:conll_luke}
     \end{subfigure}
     \hfill
     \begin{subfigure}{0.245\textwidth}
         \centering
         \includegraphics[width=\textwidth]{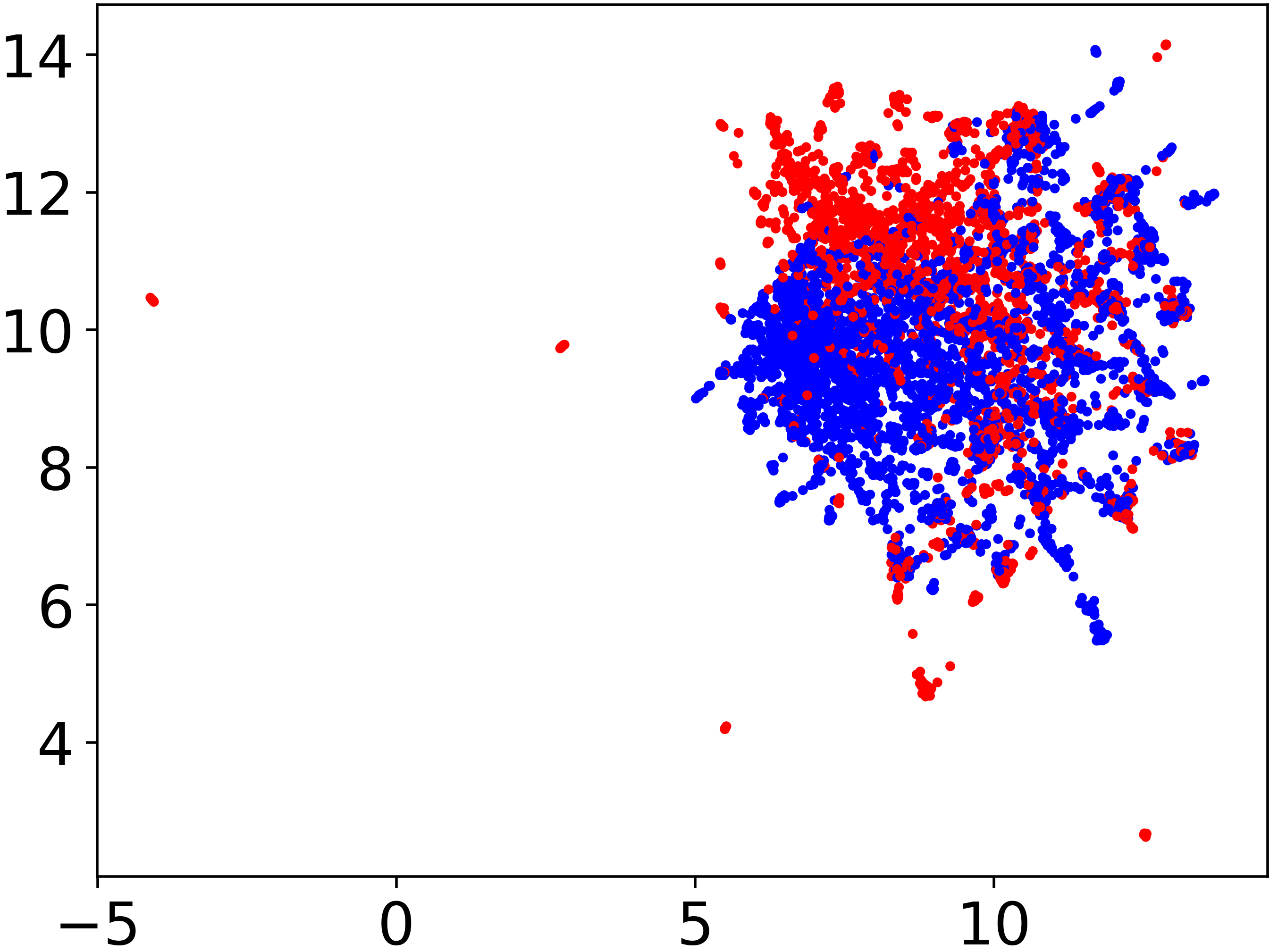}
         \caption{RoBERTa-endpoint}
         \label{fig:conll_endpoint}
     \end{subfigure}
     \hfill
     \begin{subfigure}{0.245\textwidth}
         \centering
         \includegraphics[width=\textwidth]{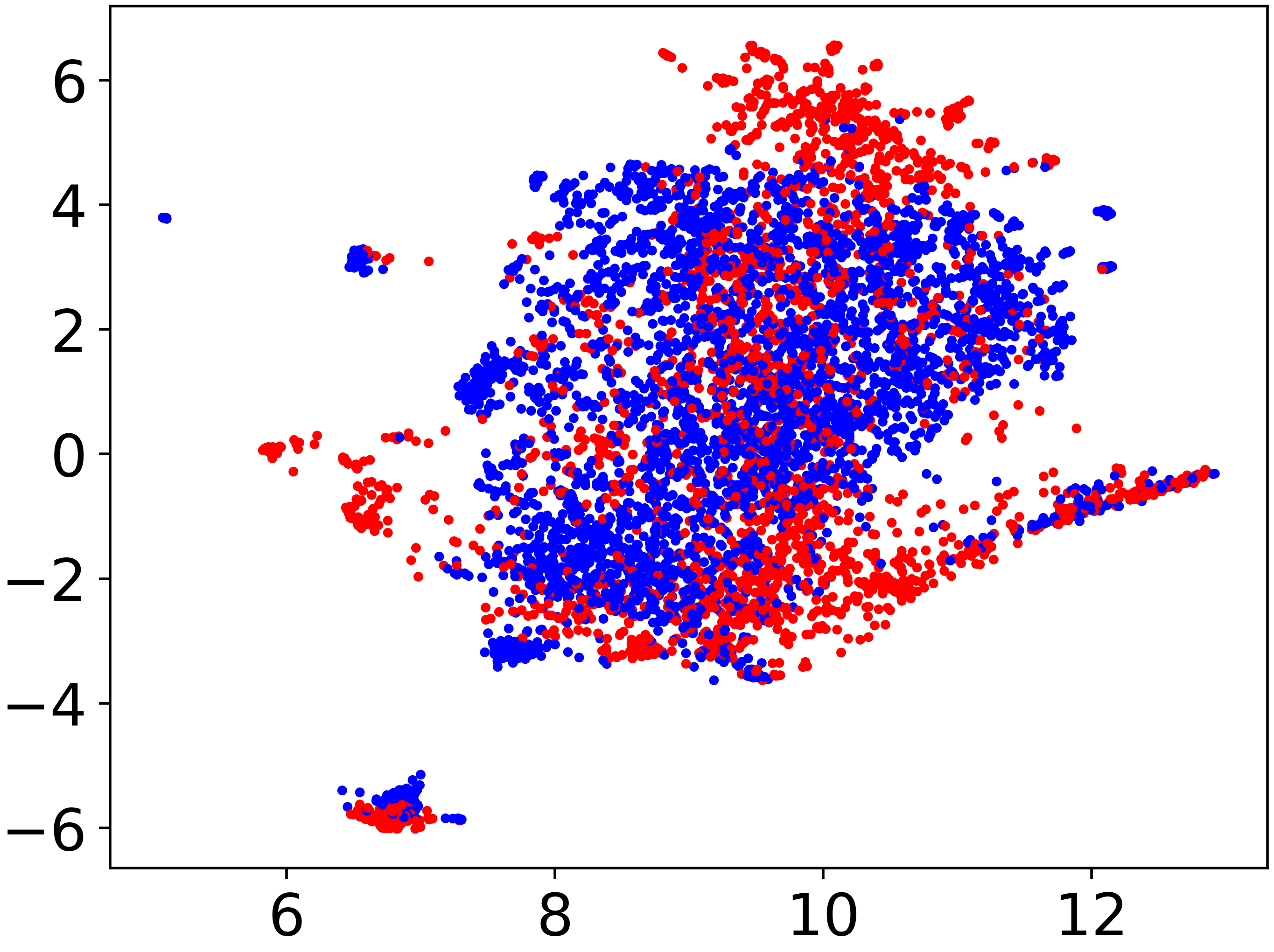}
         \caption{RoBERTa-maxpool}
         \label{fig:conll_maxpooling}
     \end{subfigure}
     }
    \vspace{-2mm}
    \caption{Embedding of Chemical (green) and Disease (yellow) entities in BC5CDR (upper line) and PER (blue) and ORG (red) entities in CoNLL2003 (lower line) from four models 
    % kinds 
    of entity representations.}
    %: (a)(e) our pre-trained entity representations, (b)(f) LUKE pre-trained entity representations, (c)(g) EndPoint concatenation of RoBERTa, and (d)(h) token MaxPooling of RoBERTa.}
    \vspace{-2mm}
    \label{fig:entity_repr}
\end{figure*}
\begin{figure*}[t]
\scalebox{0.9}{
     \centering
     \begin{subfigure}{0.245\textwidth}
         \centering
         \includegraphics[width=\textwidth]{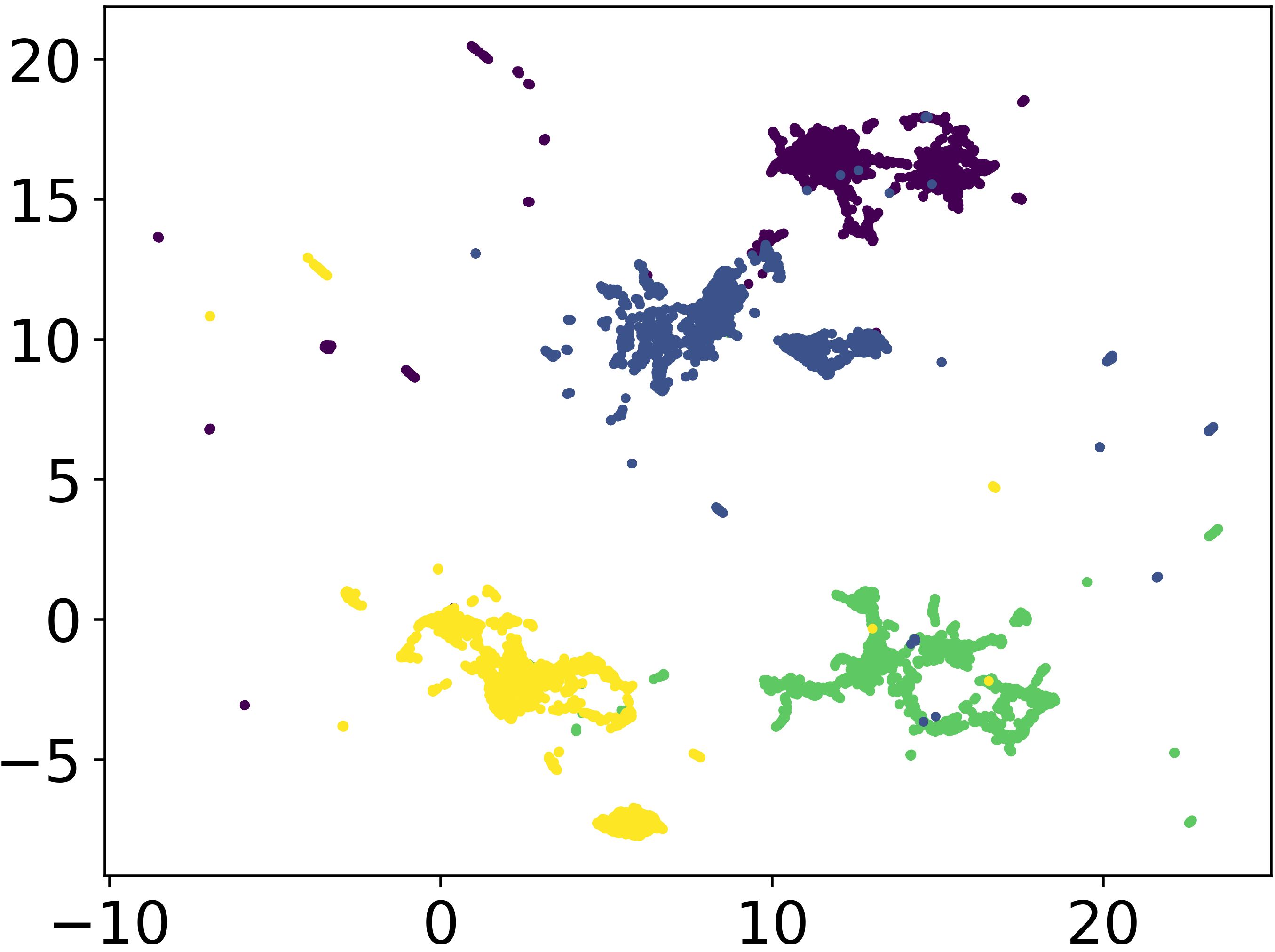}
         \caption{\textbf{SPOT}}
         \label{fig:nyt24}
     \end{subfigure}
     \hfill
     \begin{subfigure}{0.245\textwidth}
         \centering
         \includegraphics[width=\textwidth]{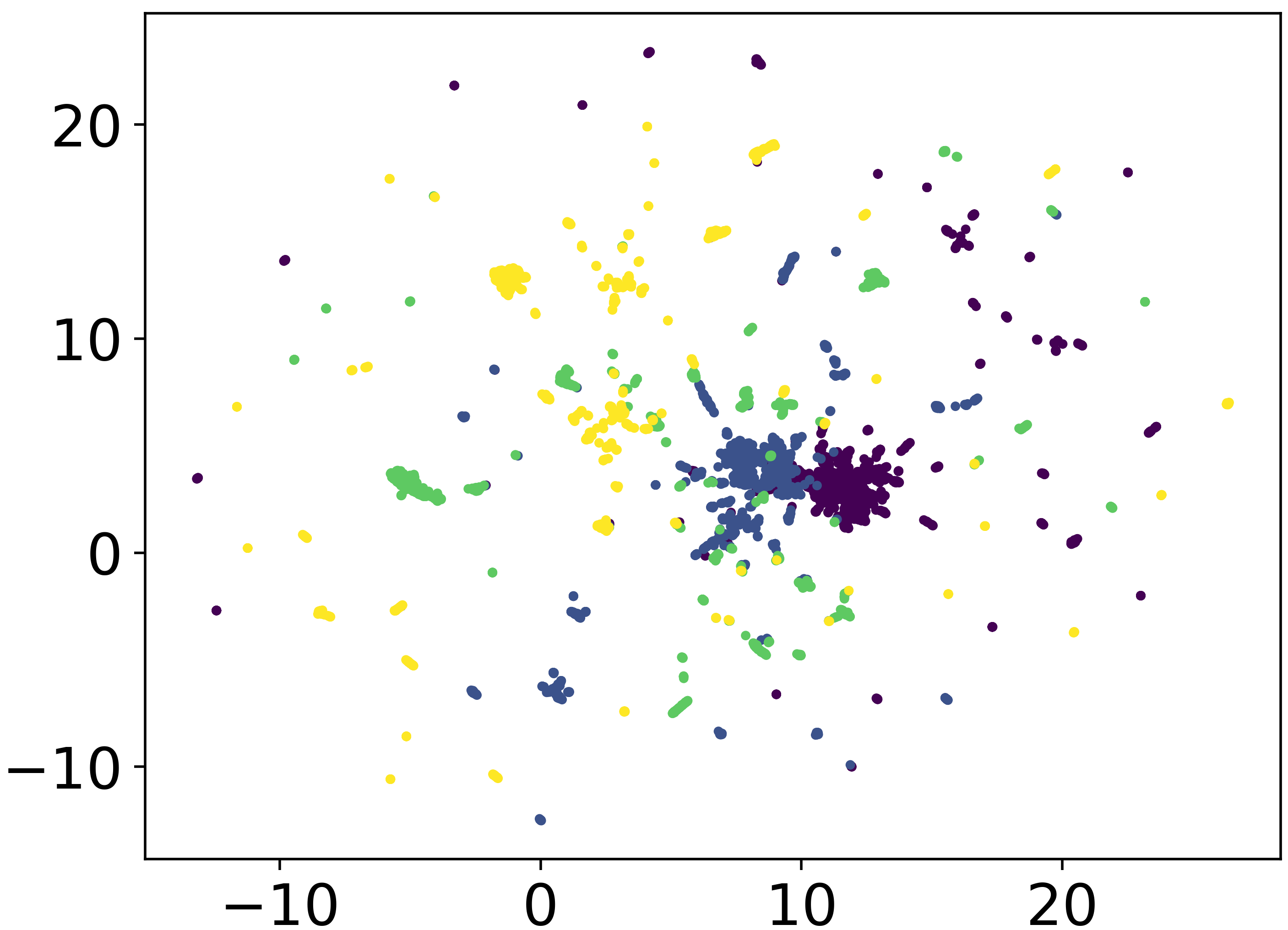}
         \caption{LUKE}
         \label{fig:nyt24_luke}
     \end{subfigure}
     \hfill
     \begin{subfigure}{0.245\textwidth}
         \centering
         \includegraphics[width=\textwidth]{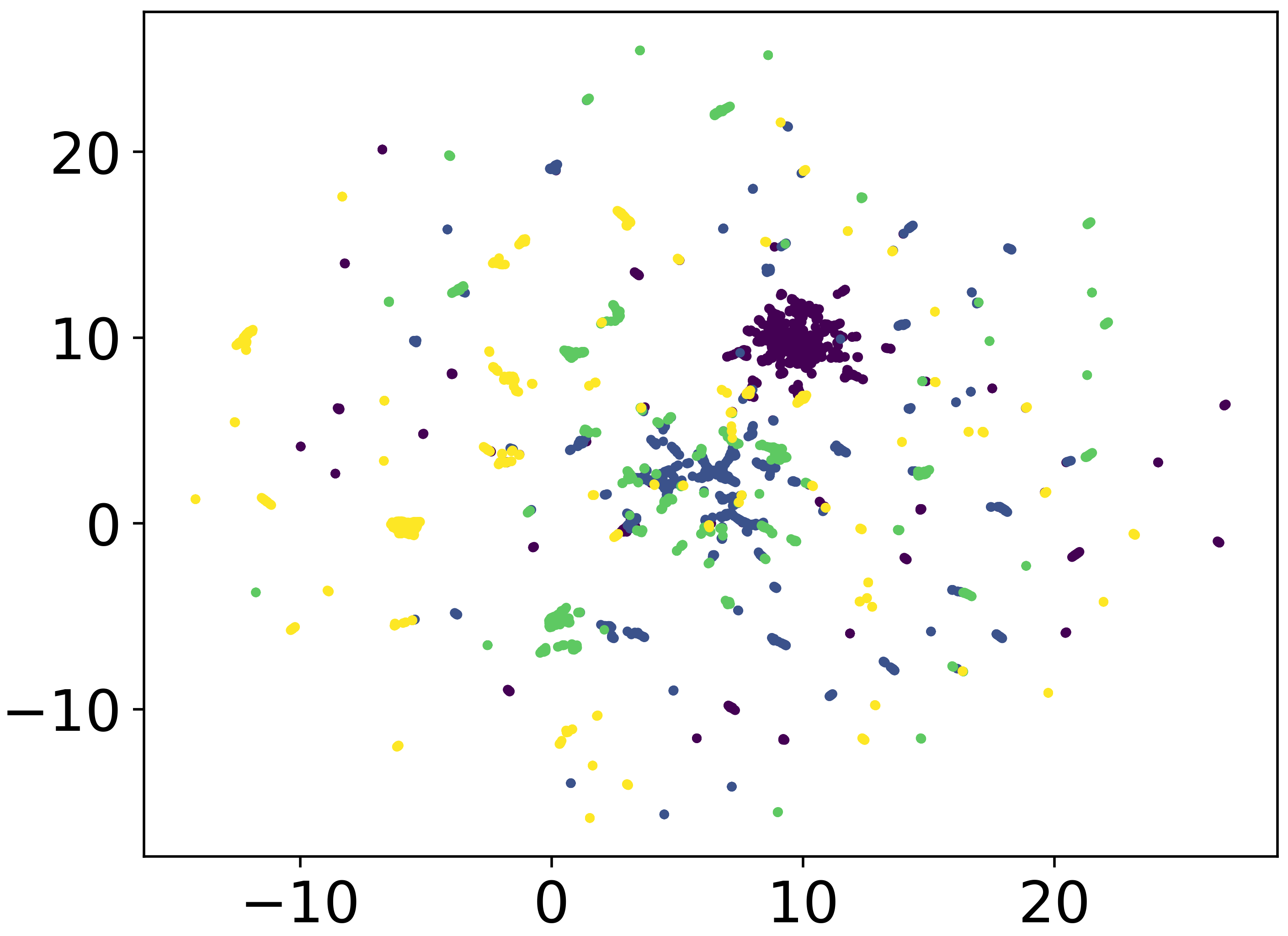}
         \caption{RoBERTa-endpoint}
         \label{fig:nyt24_endpoint}
     \end{subfigure}
     \hfill
     \begin{subfigure}{0.245\textwidth}
         \centering
         \includegraphics[width=\textwidth]{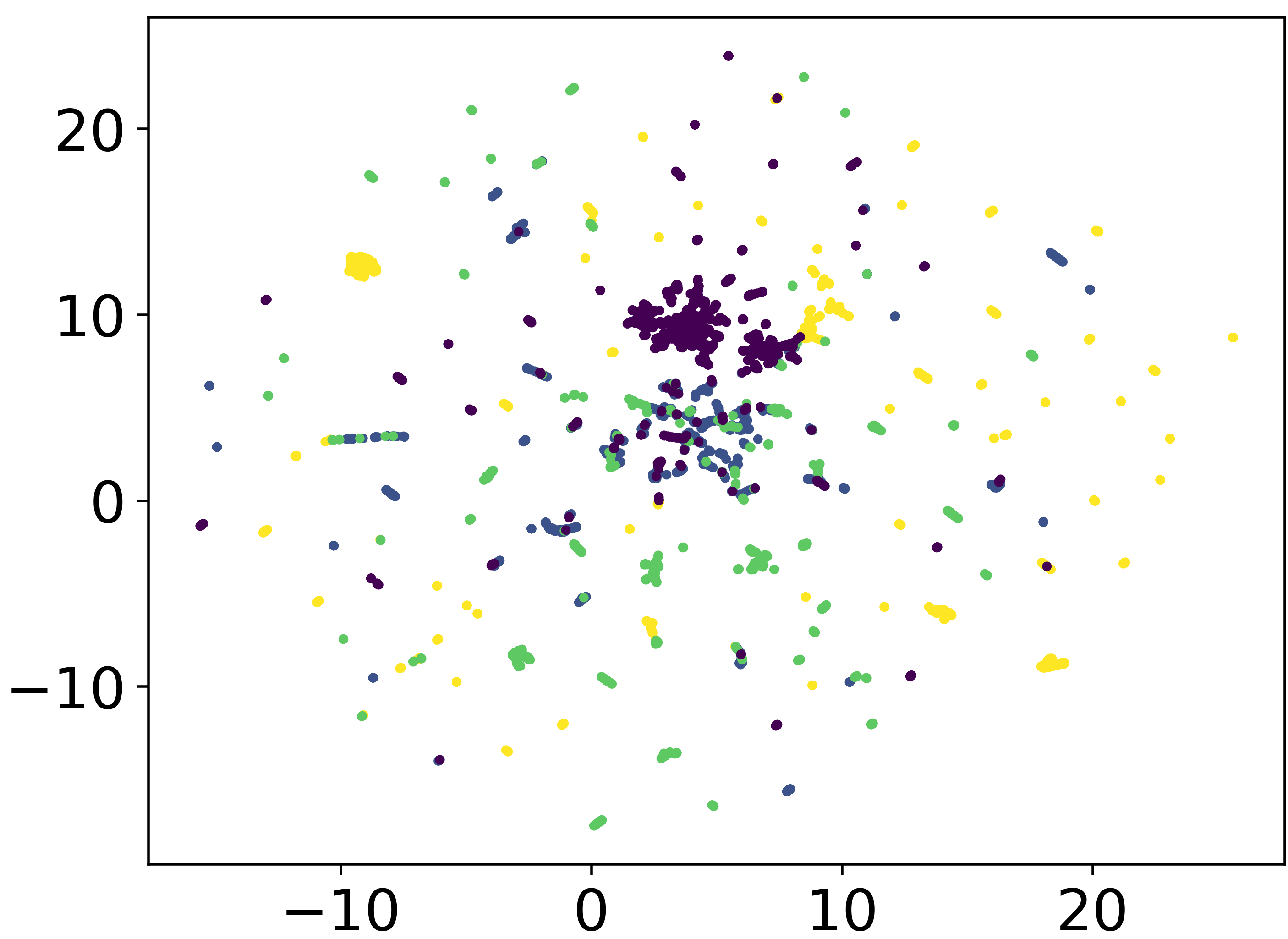}
         \caption{RoBERTa-maxpool}
         \label{fig:nyt24_maxpooling}
     \end{subfigure}
     }
     \vspace{-2mm}
    \caption{Embedding of four relations in NYT24 - place\_lived (purple), nationality (blue), country (green), and capital (yellow) - from four % kinds 
    models of relation representations.} %: (a) Our pre-trained relation representations, (b) Concatenation of entity representations from LUKE, (c) Endpoint concatenation of RoBERTa, and (d) Token maxpooling of RoBERTa.}
    \vspace{-4mm}
    \label{fig:relation_repr}
\end{figure*}

\subsection{Representation Visualization}
This section reports our study on whether our pre-trained span encoder and pair encoder can output meaningful entity and relationship representations without fine-tuning on task-specific datasets. 
\subsubsection{Entity Representations}

To show \our~can output meaningful entity representations without fine-tuning on downstream tasks, we applied pre-trained \our on BC5CDR and CoNLL2003 datasets % and 
to predict entity representations for annotated entities in the datasets. If an entity appears multiple times in the dataset, The mean of span representations was used as 
the 
entity representation. UMAP~\cite{McInnes2018UMAPUM} was adopted
%technique
to reduce the dimension of entity vectors. % In UMAP, 
The cosine similarity was the metric of % computing 
distance between vectors % is cosine similarity 
and the number of neighbors was set to $10$ for all experiments. The results were shown in Figure~\ref{fig:bc5cdr} and %Figure~
\ref{fig:conll}. We can see that \our % can 
distinguished chemical and disease 
entities well 
%and there is 
with
an obvious gap between two clusters. Compared to BC5CDR, it is more difficult for pre-trained models to separate person names and organizations in CoNLL2003 because words in these two categories are low-frequency with many overlaps.
% words in pre-trained langauge models. 
However, \our can still group entities of names or organizations together.

%\textbf{Comparison with RoBERTa.} 
To obtain entity representations without fine-tuning from RoBERTa, we adopted two methods:
\begin{enumerate}
    \item Concatenating the representations of the first and the last tokens in a span. This method is denoted as EndPoint;
    \item Maxpooling the representations of all tokens in a span.
\end{enumerate}
%The results of EndPoint on two datasets are shown in Figure~\ref{fig:bc5cdr_endpoint} and~\ref{fig:conll_endpoint} and Maxpooling's results are shown in Figure~\ref{fig:bc5cdr_maxpooling} and~\ref{fig:conll_maxpooling}. 
The figures show that RoBERTa cannot separate different types of entities well especially for low-frequency entities such as person names and organizations in CoNLL2003.

% \textbf{Comparison with LUKE.} % This is a powerful pre-trained model which can generate contextual entity representations from sentences. After obtaining span representations in each sentence, we use the same method as \our to compute entity representations and project them to 2-dimension plane. 
The results (Figure~\ref{fig:bc5cdr_luke} and~\ref{fig:conll_luke}) show that LUKE achieved similar results to \our. However, note that LUKE requires $128$M parameters for $500$K entities compared to $16$M for $1$M entities with \our. 
% with \our and distinguishes entity types but entity representations from \our are more concentrated than LUKE.

\subsubsection{Relation Representations}
Previous pre-trained models with knowledge focus on how to infuse knowledge into models~\cite{Zhang2019ERNIEEL, Peters2019KnowledgeEC, Xiong2020PretrainedEW} and LUKE~\cite{Yamada2020LUKEDC} can output entity representations, but few can 
% they rarely 
represent relationships 
% between entities 
without fine-tuning. % In this paper, 
In contrast, \our uses its 
pre-trained span pair module % introduced in 
to output relation representations.
%(Section~\ref{sec:pair}).
Again, we used the same vector projection method (i.e.,~
UMAP) with entity representations 
and 
% select 
considered four relations in NYT24:
\begin{enumerate}
    \item~\texttt{place\_lived} (purple) between person and place;
    \item~\texttt{nationality} (blue) between person and country;
    \item~\texttt{country} (green) between country and place;
    \item~\texttt{capital} (yellow) between place and country.
\end{enumerate}
%Result of \our is 
As shown in Figure~\ref{fig:nyt24}, \our can separate the four different % kinds of 
relations % and the result has 
with four clearly distinguishable clusters.

% \textbf{Comparison with RoBERTa} 
Recall that EndPoint and Maxpooling were adopted to obtain entity representations from RoBERTa. To represent relations from RoBERTa, head and tail entities were concatenated to construct their relation representations. Figures~\ref{fig:nyt24_endpoint} and~\ref{fig:nyt24_maxpooling} show that neither method can represent relations
% these two methods 
without fine-tuning on this task-specific 
dataset. Only % relation 
\texttt{place\_lived} (purple) groups together while the other three % relations 
scatter.

We concatenated~\footnote{Same as the relation representations in LUKE} contextual entity representations from LUKE to represent relations. % and result is shown in
Figure~\ref{fig:nyt24_luke} shows that
\texttt{country} green and \texttt{capital} yellow points scatter, 
%which means 
suggesting that
%LUKE cannot handle with relations \texttt{country} (green) and \texttt{capital} (yellow). C
%concatenating well pre-trained entity representations by 
LUKE cannot represent these relations well without fine-tuning.

In summary, without fine-tuning, pre-trained language models (e.g.,~RoBERTa) cannot represent entities and relations by concatenation or maxpooling. \our outputs meaningful entity and relation representations without fine-tuning

\subsection{Ablation Study}
% of Span and Pair Encoders}
\label{sec:ablation}
\begin{table}[t]
\caption{Results of ablation study. The last column % is 
shows relation results using self-attention for spans.}
\vspace{-2mm}
\centering
\begin{tabular}{@{}l|c|cc@{}}
\toprule
           & Entity & Relation & +Span attn \\ \midrule
EndPoint   & 94.92  & 76.99    & 77.48   \\ 
Self-attention  & 73.92  & 55.80     & -      \\ 
MaxPooling & 94.06  & 76.66    & -      \\ 
End+Att    & \textbf{95.05}  & 78.80     & \textbf{79.5}   \\ 
End+Max    & 94.59  & 78.23    & -      \\ 
Att+Max    & 87.27  & 62.08    & -      \\ \bottomrule
\end{tabular}
\vspace{-4mm}
\label{tab:ablation}
\end{table}
Recall that in %the \emph{Method} section 
Section~\ref{sec:model_arch},
we introduced EndPoint, Self-attention, and Maxpooling as 
%the 
options for \our's
span encoder and proposed to use span attention to obtain contextual span representations. 
% In this section, we will discuss the best span encoder and pair encoders from these methods. 
We conducted experiments with BERT on the NYT24 dataset and compare micro F$_1$ scores of NER and relation extraction results
to validate our choice. We set the same hyper-parameters for all experiments. Specifically, the batch size was $16$ and Adam optimizer was adopted with learning rate 2e-5 and linear learning rate decay. All gradients were clipped to $5$ and warm up steps were $300$. We adopted early stop by using the best model on development set and tested on a test set. When we used two kinds of encoders, we concatenated the two outputs of the encoders and fed the concatenation to a linear classifier.

% From 
Table~\ref{tab:ablation} suggests %, we can see 
that the most important method for span representation is EndPoint plus % and 
Self-attention (End+Att).
% which
Self-attention is known as 
% cannot solely 
ineffective when used solely
as a span encoder but here we show that it 
% Self-attention 
can improve relation representations
% results 
when working with EndPoint. Contextual span representations by span attention improve relation results because more contextual information can be % is 
learned which is important for relation extraction. % task. 

%% file: 5_conclusion.tex
In this paper, we proposed \our~to represent knowledge by spans and described a method to incorporate knowledge information into a language representation model. Accordingly, we proposed a span encoder and a span pair encoder to represent knowledge (i.e.,~entities and relationships in a knowledge graph) in text. \our~can represent knowledge by spans
without extra entity embeddings and entity lookup, % by spans. 
requiring a much less number of parameters than existing 
knowledge-enhanced language models attempting to represent knowledge in language models.
Our experimental results demonstrated that \our 
generated high-quality entity and relation representations based on our clustering experiments and achieved superior results on three supervised information extraction tasks. 
Overall, the results show that \our was more effective in 
representing entities and relationships without fine-tuning
while requiring an order of magnitude less parameters
than previous methods.